\documentclass{article}

\PassOptionsToPackage{numbers, compress}{natbib}

\usepackage[preprint]{neurips_2023}

\usepackage[utf8]{inputenc} 
\usepackage[T1]{fontenc}    
\usepackage{hyperref}       
\usepackage{url}            
\usepackage{booktabs}       
\usepackage{amsfonts}       
\usepackage{nicefrac}       
\usepackage{microtype}      
\usepackage{xcolor}         
\usepackage{bbm}
\usepackage{bm}
\usepackage[noend]{algpseudocode}

\usepackage{algorithmicx,algorithm,amsmath}
\usepackage{amsthm}

\usepackage{appendix}
\usepackage{threeparttable}
\usepackage{multirow}
\usepackage{utfsym}
\usepackage{graphicx}
\usepackage{subcaption}
\usepackage{epstopdf}
\usepackage{wrapfig}
\usepackage{amssymb}
\allowdisplaybreaks[4]

\usepackage{color, colortbl}
\definecolor{Gray}{gray}{0.9}

\usepackage{ulem} 

\usepackage{soul}
\soulregister{\cite}7 
\soulregister{\citep}7 
\soulregister{\citet}7 
\soulregister{\ref}7 
\soulregister{\pageref}7 

\title{Privacy-Preserving Federated Learning with Consistency via Knowledge Distillation Using Conditional Generator}

\author{Kangyang Luo$^{1}$, Shuai Wang$^{1}$,  Xiang Li$^{1}$\thanks{Corresponding author}, Yunshi Lan$^{1}$, Ming Gao$^{1}$, Jinlong Shu$^{2}$\\
{\tt\small East China Normal University$^{1}$, Shanghai Normal University$^{2}$}\\
{\tt\small Shanghai, China}\\
{\tt\small \{52205901003, 51215903042\}@stu.ecnu.edu.cn,} \\ 
{\tt\small \{xiangli, yslan, mgao\}@dase.ecnu.edu.cn, jlshu@shnu.edu.cn}
}

\begin{document}

\maketitle

\begin{abstract}
Federated Learning~(FL) is gaining popularity as a distributed learning framework that only shares model parameters or gradient updates and keeps private data locally.
However, FL is at risk of privacy leakage caused by privacy inference attacks. 
And most existing privacy-preserving mechanisms in FL conflict with achieving high performance and efficiency. 
Therefore, we propose FedMD-CG, a novel FL method with highly 
competitive performance and high-level privacy preservation, which decouples each client's local model into a feature extractor and a classifier, and utilizes a conditional generator instead of the feature extractor to perform server-side model aggregation.
To ensure the consistency of local generators and classifiers, FedMD-CG leverages knowledge distillation to train local models and generators at both the latent feature level and the logit level.
Also, we construct additional classification losses and design new diversity losses to enhance client-side training.
FedMD-CG is robust to data heterogeneity and does not require training extra discriminators~(like cGAN).
We conduct extensive experiments on various image classification tasks to validate the superiority of FedMD-CG. 
We provide our code here:
\href{https://anonymous.4open.science/r/FedMD-CG-34E2/}{https://anonymous.4open.science/r/FedMD-CG-34E2/}.
\end{abstract}

\section{Introduction}
Many modern real-world applications involve data being dispersed across clients located in different physical locations, such as autonomous driving~\cite{Li2021Privacy}, medical image analysis~\cite{Liu2021Feddg}, and IoT~\cite{Nguyen2021Federated}.
However, various regulation, 
privacy and security concerns often make it impractical or even impossible to collect these dispersed data into one location for traditional centralized learning~\cite{Voigt2017eu}.
To ameliorate these limitations, Federated Learning~(FL)~\cite{Li2020Federated1} has been proposed to enable each client to train a local model using only its own data and share its model parameters or gradient updates with a central server periodically to ensure that each client's raw data does not leak from the local device.

Despite the success, 
the vanilla FL~(e.g., FedAvg~\cite{McMahan2017Communication} and its variants~\cite{Li2020Federated, Karimireddy2020Scaffold, Luo2023GradMA, luo2023dfrd, luo2024dfdg}) based on sharing complete local model parameters or gradient updates are extremely vulnerable to inference attacks.
Several prior arts empirically demonstrate that it is feasible to reconstruct victim clients' private data from trained and publicly shared parameters and gradient updates~\cite{Zhu2019Deep, Geiping2020Inverting, Haim2022Reconstructing}.
Therefore, a variety of efforts have been devoted to 
reducing 
the risk of privacy leakage in FL, including homomorphic encryption~(HE)~\cite{Ma2022Privacy, Zhang2022Homomorphic}, differential privacy~(DP)~\cite{Geyer2017Differentially, Cheng2022Differentially} and model decoupling~(MD)~\cite{Arivazhagan2019Federated, Liang2020Think}.
In particular,
HE achieves high-level privacy protection at the expense of extremely high computation and communication costs, which restricts its deployment in bandwidth-limited and large-model scenarios.
DP preserves privacy by perturbing server-side model aggregation or client-side local model update, but this deteriorates the performance of the FL methods. 
See Appendix~\ref{Related Work:} for more related work.

In this paper, we mainly focus on MD, which requires each client to decompose the local model into the base and top layers, and send one of them to the server to reduce the risk of privacy leakage, 
yet this inevitably 
results in performance degradation and even privacy exposure.
Note that we regard the base layers~(top layers) as a feature extractor~(classifier).
Recently, FedCG~\cite{Wu2021Fedcg} combines FL and conditional generative adversarial network~(cGAN)~\cite{Mirza2014Conditional} to adversarially train a conditional generator to replace the feature extractor for each client, aiming at achieving competitive performance while maintaining high-level privacy protection.
However, we revisit it and observe that the following pitfalls may occur in client-side training. 
First, knowledge transfer modality at the latent feature level may not be sufficient. 
Second, additional discriminators need to be trained to satisfy the adversarial training of cGAN. 
Third, the trained local generator may not match the local classifier, terming their inconsistency. 
Note that the latent feature denotes the output of the feature extractor.

\begin{figure}[!t]
\centering
\includegraphics[width=0.9\textwidth]{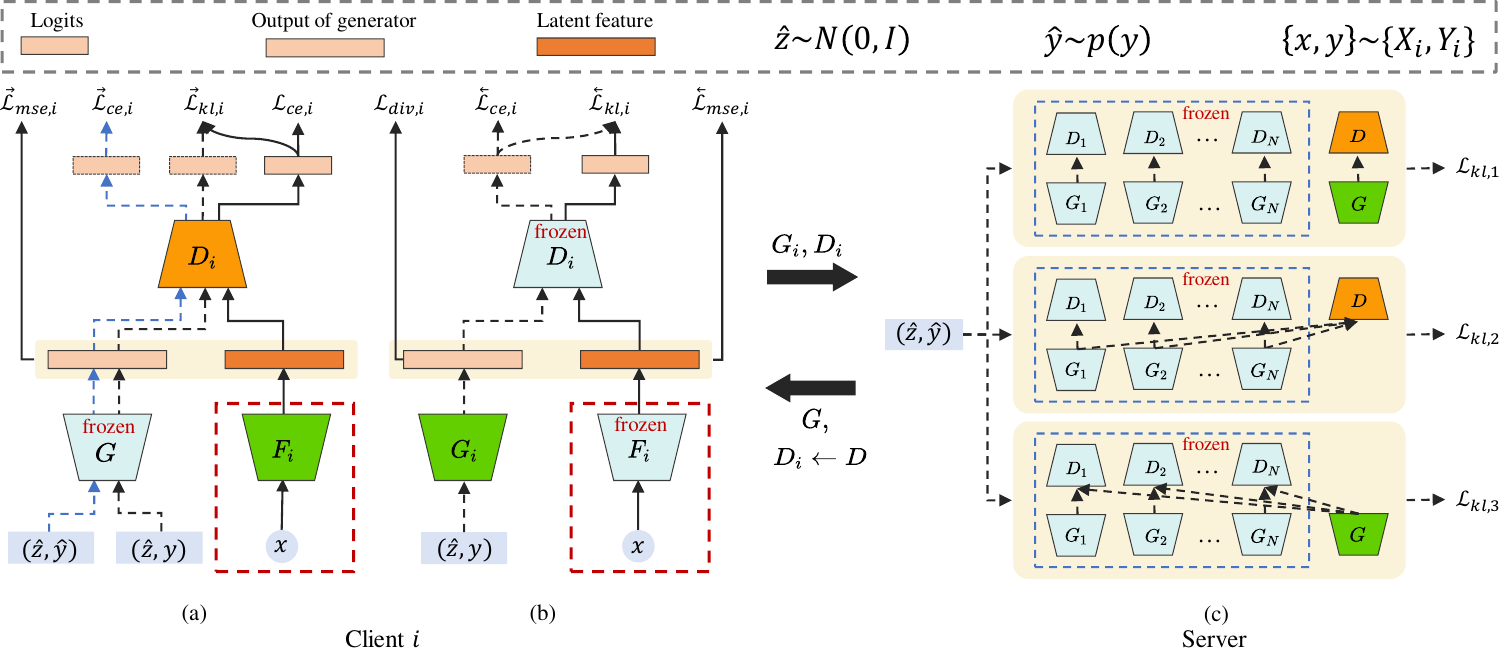} 
\caption{Illustration of FedMD-CG: (a) The local model update distills the experience from the global generator $G$ for augmenting the generalization 
performance of the local model [$F_i$, $D_i$]. 
(b) The local generator update utilizes the trained local model [$F_i$, $D_i$] to guide the local generator $G_i$ to mimic latent feature space. 
Note that $G$ is not involved in client-side training.
(c) The server-side data-free KD aggregation takes a crossed manner to achieve as much knowledge transfer as possible. 
Best viewed in color. Zoom in for details.}
\label{pic2:}
\vspace*{-2ex}
\end{figure}

To this end, we propose a new \underline{\textbf{Fed}}erated Learning with \underline{\textbf{MD}} method~(dubbed as FedMD-CG), which  resorts to knowledge distillation~(KD) to train a local \underline{\textbf{c}}onditional \underline{\textbf{g}}enerator for each client to replace the local feature extractor. 
To be more specific, FedMD-CG works on how to efficiently train  the local model and generator on the client side.
To achieve this, FedMD-CG utilizes KD to perform knowledge transfer from the global generator to the local model and from the local model to the local generator at the latent feature level and the logit level.
Meanwhile, we additionally construct two classification losses to enhance the local model update and the local generator update, respectively.
In addition, we devise two novel diversity constraints to ensure the diversity of the local generator outputs.
On the server side, FedMD-CG performs aggregation of local generators and classifiers in a crossed data-free KD fashion.
The overview of our method is illustrated in Fig.~\ref{pic2:}. 

In a nutshell, the main contributions of this work are as follows: 
\begin{itemize} 
    \item  We formulate a novel privacy-preserving FL method FedMD-CG to achieve better generalization performance, via leveraging KD to efficiently transfer knowledge from the global generator to the local model and then from the local model to the local generator.
    
    \item To enhance client-side training, we construct additional classification losses and tailor new diversity constraints. 
    Our method ensures the consistency between trained local generators and classifiers, thereby being robust to data heterogeneity.

    \item FedMD-CG performs aggregation in a crossed data-free KD fashion on the server side in order to extract as much knowledge as possible from the local generators and classifiers.

    \item We conduct extensive experiments to show that FedMD-CG
    is highly competitive compared with state-of-the-art baselines w.r.t test performance, convergence speed and privacy protection.
\end{itemize}

\section{Proposed Method} 
In this section, we detail the proposed method FedMD-CG. We first define the problem setup and notations for clarity.
And then the core modules of FedMD-CG are presented. Moreover, we present pseudocode for FedMD-CG in Appendix~\ref{sec:pseudo}.

\textbf{Problem Setup and Notations.} In this work, we consider supervised federated learning~(FL) setting, i.e., the general problem of multi-class classification.
To be specific, we focus on 
the centralized setup that consists of a central server and $N$ clients owning private labeled datasets $\{(\bm{X}_i, \bm{Y}_i)\}_{i=1}^N$ with $|\bm{X}_i|=n_i$, where $\bm{X}_i=\{\bm{x}_i^b\}_{b=1}^{n_i}$ follows data distribution $\mathcal{D}_i $ over input feature space $\mathcal{X}_i$, i.e., $\bm{x}_i^b \sim \mathcal{D}_i$, and $\bm{Y}_i=\{y_i^b\}_{b=1}^{n_i} \subset \mathbbm{R}$ denotes the ground-truth labels of $\bm{X}_i$.
Notably, 
we consider the same input feature space, 
yet the sample distribution may be different among clients, that is, data heterogeneity caused by label distribution skewness~(i.e., $\mathcal{X}_i=\mathcal{X}_j$ and $\mathcal{D}_i \neq \mathcal{D}_j, \forall i\neq j, i,j \in [N]$). %
Besides, each client $i$ holds a local model parameterized by $\bm{\theta}^i=[\bm{\theta}_F^i; \bm{\theta}_D^i]$ comprising 
two components: the base layers~(feature extractor) 
$F_i: \bm{X}_i \rightarrow \mathcal{F}$ parameterized by $\bm{\theta}_F^i$, and the top layers~(classifier) 
$D_i: \mathcal{F} \rightarrow \mathbbm{R}^c$ parameterized by $\bm{\theta}_D^i$, where $\mathcal{F} \subset \mathbbm{R}^p$ is the output space of feature extractor with $p$ dimension, i.e., the latent feature space, and $c$ is the number of classes. 
For extracting knowledge from clients without accessing any extra data, each client equips with a conditional generator $G: \mathcal{Z}\times \mathcal{Y}\rightarrow \mathbbm{R}^p$ parameterized by $\bm{w}$, where $\mathcal{Z} \subset \mathbbm{R}^q$ is the multivariate standard normal distribution $\mathcal{N} (\bm{0}, \bm{I})$ and $\mathcal{Y} \subset \mathbbm{R}^c$ indicates the one-hot vector space of the ground-truth label. 
We use bold $\bm{y} \in \mathcal{Y}$ to denote one-hot vector corresponding to class $y \in \mathbbm{R}$. 
Hereafter, we refer to conditional generator as generator.

\subsection{Client-side Two-stage Distillation}
The training process for each client $i$ involves two stages: augmenting the local model update with global generator~(see Fig.~\ref{pic2:}~(a)), and guiding the local generator update with trained local model~(see Fig.~\ref{pic2:}~(b)).

\textbf{Augmenting the local model update with global generator.} 
In the classical local model update,
client $i$ leverages the following classification loss to optimize the local model $\bm{\theta}^i=[\bm{\theta}^i_F, \bm{\theta}^i_D]$:
\begin{align}
     \mathcal{L}_{ce,i}=CE(\rho(D_i(F_i(\bm{x}))), y), 
\end{align}
where $\rho$ is the softmax function and $CE$ is the cross-entropy function. 
However, $\mathcal{L}_{ce,i}$ has no access to global knowledge in our work, which is embedded in the global generator. To transfer the knowledge of the global generator to the local model efficiently, we construct the following two losses based on KD:
\begin{align}
     &{\mathop{\mathcal{L}}\limits ^{\to}}_{mse, i}=\|F_i(\bm{x})-G(\hat{\bm{z}},\bm{y})\|^2,  {\mathop{\mathcal{L}}\limits ^{\to}}_{kl, i}=KL(\rho(D_i(F_i(\bm{x})))\|\rho(D_i(G(\hat{\bm{z}},\bm{y})))),
\end{align}
where $\hat{\bm{z}}$ is sampled from $\mathcal{N}(\bm{0}, \bm{I})$. $\|\cdot\|^2$ and $KL$ are $L_2$-norm function and Kullback-Leibler function, respectively.
Specifically,
${\mathop{\mathcal{L}}\limits ^{\to}}_{mse, i}$ utilizes $L_2$-norm function to enforce the output of feature extractor~$F_i(\bm{x})$ to approximate that of global generator $G(\hat{\bm{z}}, \bm{y})$. 
After that, 
client $i$ feeds both $F_i(\bm{x})$ and $G(\hat{\bm{z}}, \bm{y})$ into 
the classifier to get $D_i(F_i(\bm{x}))$ and $D_i(G(\hat{\bm{z}},\bm{y}))$.
Further, ${\mathop{\mathcal{L}}\limits ^{\to}}_{kl, i}$ harnesses Kullback-Leibler function to make $D_i(F_i(\bm{x}))$ close to $D_i(G(\hat{\bm{z}},\bm{y}))$. 

To further augment the local model update, client $i$ resamples a batch of noisy data to feed the generator and classifier sequentially, and minimizes the following classification loss:
\begin{align}
     &{\mathop{\mathcal{L}}\limits ^{\to}}_{ce, i}=CE(\rho(D_i(G(\hat{\bm{z}}, \hat{\bm{y}}))), \hat{y}),
\end{align}
where $\hat{y} \sim p(y) \propto \sum_{i\in [N]}n_i^y$, $n_i^y$ denotes the number of samples w.r.t class $y$ on the $i$-th client.

Combining $\mathcal{L}_{ce,i}$, $ {\mathop{\mathcal{L}}\limits ^{\to}}_{ce, i}$, ${\mathop{\mathcal{L}}\limits ^{\to}}_{mse, i}$ and ${\mathop{\mathcal{L}}\limits ^{\to}}_{kl, i}$, the overall objective of the local model update can be formalized as follows:
\begin{align}
     \label{gen_t_local:}
     \min_{\bm{\theta}^i_F, \bm{\theta}^i_D}&\mathbbm{E}_{\substack{\hat{\bm{z}}, \hat{y}\sim \mathcal{N}(\bm{0},\bm{I}), p(y)\\ \bm{x}, y \sim \bm{X}_i, \bm{Y}_i}} [\mathcal{L}_{ce, i} + \lambda_{1}{\mathop{\mathcal{L}}\limits ^{\to}}_{ce, i} + 
     \lambda_{2}{\mathop{\mathcal{L}}\limits ^{\to}}_{mse, i} + \lambda_{3}{\mathop{\mathcal{L}}\limits ^{\to}}_{kl, i}],
\end{align}
where $\lambda_{1}$, $\lambda_{2}$ and $\lambda_{3}$ are tunable hyperparameters for balancing different loss items. 


\begin{figure*}[!t]
\centering
\includegraphics[width=1.0\textwidth]{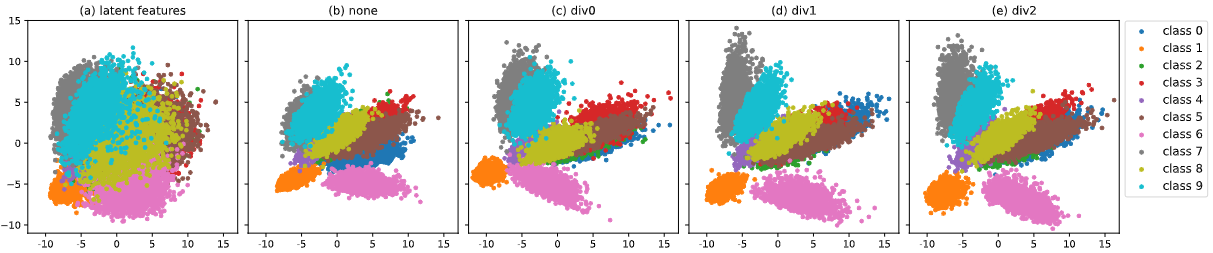} 
\vspace*{-3ex}
\caption{Visualization for output of the generator: The toy example first trains a LeNet~\cite{LeCun1998Gradient} as teacher model~(T) using the training set of MNIST~\cite{LeCun1998Gradient}. Then the test set of MNIST is fed to T to get the latent features. And the dimensions of the latent features are reduced by principal component analysis~(PCA)~\cite{Halko2011Finding}. (a) shows the latent features distribution of T after PCA dimension reduction. Next, we let T guide the training of the generator according to Eq.~(\ref{local_t_gen:}). Similarly, we utilize PCA to perform dimension reduction for the output of the generator. (b) visualizes the output distribution of the generator without  diversity constraint. 
(c), (d) and (e) visualize the output distribution of the generator with $\mathcal{L}_{div}^0$, $\mathcal{L}_{div}^1$ and $\mathcal{L}_{div}^2$, respectively.}
\label{divloss:}
\end{figure*}

\textbf{Guiding the local generator update with trained local model.}
After the local model update, we maintain a local generator in client $i$ to extract the knowledge of the trained local model without accessing its private data.
Note that the global generator does not replace the local generator in our work to learn the knowledge of the trained local model.

Similar to the manner of augmenting local model update, we utilize KD to construct losses ${\mathop{\mathcal{L}}\limits ^{\gets}}_{kl, i}$ and ${\mathop{\mathcal{L}}\limits ^{\gets}}_{mse, i}$ to transfer the knowledge of the local model to the local generator. ${\mathop{\mathcal{L}}\limits ^{\gets}}_{kl, i}$ and ${\mathop{\mathcal{L}}\limits ^{\gets}}_{mse, i}$ take the forms:
\begin{align}
     & {\mathop{\mathcal{L}}\limits ^{\gets}}_{mse, i}=\|G_i(\hat{\bm{z}},\bm{y})-F_i(\bm{x})\|^2,  {\mathop{\mathcal{L}}\limits ^{\gets}}_{kl, i}=KL(\rho(D_i(G_i(\hat{\bm{z}},\bm{y})))\|\rho(D_i(F_i(\bm{x})))).
\end{align}

To ensure the fidelity of the output of the local generator $G_i$, $G_i$ is expected to fit the input space of the local classifier for better knowledge extraction from the local model. Therefore, client $i$ takes the following classification loss ${\mathop{\mathcal{L}}\limits ^{\gets}}_{ce, i}$ to enforce $G_i$ to yield higher prediction on class $y$:

\begin{align}
     {\mathop{\mathcal{L}}\limits ^{\gets}}_{ce, i}=CE(\rho(D_i(G_i(\hat{\bm{z}}, \bm{y}))), y).
\end{align}

However, if we only optimize ${\mathop{\mathcal{L}}\limits ^{\gets}}_{kl, i}$, ${\mathop{\mathcal{L}}\limits ^{\gets}}_{mse, i}$ and ${\mathop{\mathcal{L}}\limits ^{\gets}}_{ce, i}$ for $G_i$, it is likely to generate similar outputs for each class with little diversity, which can cause the model collapse of the local generator. 
To tackle this limitation, the constraint $\mathcal{L}_{div}^0$ has been added to enhance the output diversity of the generator as follows~\cite{Yoo2019Knowledge, Zhu2021Data, Zhang2022Fine}:
\begin{align}
     \label{div_0:}
     \mathcal{L}_{div}^0=e^{\frac{1}{B^2}\sum\limits_{j,k\in [B]}\left(-\|\hat{\bm{f}}_j-\hat{\bm{f}}_k\|_2*\|\hat{\bm{z}}_j-\hat{\bm{z}}_k\|_2\right)},
\end{align}
where $B$ denotes the batch size and $\hat{\bm{f}}_{j/k}=G_i(\hat{\bm{z}}_{j/k}, \bm{y}_{j/k})$.
This constraint treats the noise pair distance $\|\hat{\bm{z}}_j-\hat{\bm{z}}_k\|_2$ as a weight, which is then multiplied by the corresponding output pair distance $\|\hat{\bm{f}}_j-\hat{\bm{f}}_k\|_2$ in each batch $B$, thus imposing more weights on the output pairs whose corresponding noise pairs are more distant. 
It can be found that this weighting scheme of $\mathcal{L}_{div}^0$ is label-agnostic.
In other words, the weight differences between intra- and inter-class output pairs are not considered in $\mathcal{L}_{div}^0$, which may lead to inter-class output pairs being close but intra-class output pairs being distant, thus adversely affecting the performance of $G_i$.
To rectify this issue, we propose two novel 
diversity constraints, $\mathcal{L}_{div}^1$ and $\mathcal{L}_{div}^2$.
In terms of $\mathcal{L}_{div}^1$, we simply replace $\|\hat{\bm{z}}_j-\hat{\bm{z}}_k\|_2$ of Eq.~(\ref{div_0:}) with $\|\hat{\bm{z}}_j^y-\hat{\bm{z}}_k^y\|_2$, where $\hat{\bm{z}}_j^y=[\hat{\bm{z}}_j; \bm{y}_j]$. 
Further, we formulate $\mathcal{L}_{div}^2$ in the following form:
\begin{align}
     \mathcal{L}_{div}^2=e^{\frac{1}{B^2}\sum\limits_{j,k\in [B]}\left(-\|\hat{\bm{f}}_j-\hat{\bm{f}}_k\|_2*\|\hat{\bm{z}}_j-\hat{\bm{z}}_k\|_2* e^{\|\hat{\bm{y}}_j-\hat{\bm{y}}_k\|_1}\right)}.
\end{align}

Compared to $\mathcal{L}_{div}^0$, $\mathcal{L}_{div}^1$ and $\mathcal{L}_{div}^2$ further differentiate the weights of the generator's intra- and inter-class output pair distances, with more weights applied to the inter-class output pair distances.
For brevity, we uniformly denote $\mathcal{L}_{div}^0$, $\mathcal{L}_{div}^1$ and $\mathcal{L}_{div}^2$ as $\mathcal{L}_{div}$ unless otherwise noted. 
In Fig.~\ref{divloss:}, we provide a toy example showing the output distribution of the generator without diversity constraints as well as with different diversity constraints.

We combine ${\mathop{\mathcal{L}}\limits ^{\gets}}_{kl, i}$, ${\mathop{\mathcal{L}}\limits ^{\gets}}_{mse, i}$, ${\mathop{\mathcal{L}}\limits ^{\gets}}_{ce, i}$ and $\mathcal{L}_{div}$ to yield the overall objective of the local generator update for client $i$ is shown below:
\begin{align} 
     \label{local_t_gen:}
     \min_{\bm{w}_i} & \mathbbm{E}_{\substack{\hat{\bm{z}}\sim \mathcal{N}(\bm{0},\bm{I})\\ \bm{x}, y \sim \bm{X}_i, \bm{Y}_i}} [{\mathop{\mathcal{L}}\limits ^{\gets}}_{kl, i} + \lambda_{4} {\mathop{\mathcal{L}}\limits ^{\gets}}_{mse, i} + \lambda_{5}  {\mathop{\mathcal{L}}\limits ^{\gets}}_{ce, i}+\lambda_{6} \mathcal{L}_{div, i}],
\end{align}
where $\lambda_{4}$, $\lambda_{5}$ and $\lambda_{6}$ are non-negative hyperparameters. $\mathcal{L}_{div, i}$ denotes the diversity constraint of client $i$. 

\subsection{Server-side Crossed Distillation Aggregation}
After gathering local generators and classifiers uploaded by clients, the server aggregates them as a preliminary global generator and classifier via weighted averaging.
However, straightforward average aggregation may counteract the local knowledge from clients. 
To alleviate this issue, we train the preliminary global generator and classifier via crossed data-free KD to distill as much knowledge as possible from the local generators and classifiers. 
Fig.~\ref{pic2:}~(c) shows the distillation schema on the server, where the overall distillation objective consists of three parts: $\mathcal{L}_{kl, 1}$, $\mathcal{L}_{kl, 2}$ and $\mathcal{L}_{kl, 3}$.

Specifically, the server first samples $(\hat{\bm{z}}, \hat{y})$, and feeds it to the local generators $\{G_i\}_{i\in[N]}$ and the global generator $G$, where $\hat{\bm{z}} \sim \mathcal{N}(\bm{0}, \bm{I})$ and $\hat{y}\sim p(y) \propto \sum_{i\in [N]}n_i^y$.
Their outputs are then fed into the corresponding classifiers to compute the loss $\mathcal{L}_{kl, 1}$:
\begin{align}
     & \mathcal{L}_{kl, 1} =  \sum_{i\in[N]} \tau_{i, \hat{y}} KL(\rho_g\|\rho_i),
\end{align}
where $\rho_g=\rho(D(G(\hat{\bm{z}}, \hat{\bm{y}})))$, $\rho_i = \rho(D_i(G_i(\hat{\bm{z}}, \hat{\bm{y}})))$, and $\tau_{i, \hat{y}} = n_i^{\hat{y}}/\sum_{ j\in [N]}n_j^{\hat{y}}$.
$\mathcal{L}_{kl, 1}$ ensures that the $\rho_g$ from the global classifier approximates $\{\rho_i\}_{i\in[N]}$ from the local classifiers.

However, simply distilling knowledge by minimizing $\mathcal{L}_{kl, 1}$ could be insufficient,
since $D_i$ fits $G_i(\hat{\bm{z}}, \hat{\bm{y}})$~(via optimizing Eq.~(\ref{local_t_gen:})) but may not fit $G(\hat{\bm{z}}, \hat{\bm{y}})$ and $G_i(\hat{\bm{z}}, \hat{\bm{y}})$ fits $D_i$ but may not fit $D$, such that only partial knowledge from clients can be extracted. Therefore, to address these limitations, we introduce a crossover strategy and formulate two losses $\mathcal{L}_{kl, 2}$ and $\mathcal{L}_{kl, 3}$ as:
\begin{align}
     & \mathcal{L}_{kl, 2} =  \sum_{i\in[N]} \tau_{i, \hat{y}} KL(\rho_{ig}\|\rho_i),  \mathcal{L}_{kl, 3} =  \sum_{i\in[N]} \tau_{i, \hat{y}} KL(\rho_{gi}\|\rho_i),
\end{align}
where $\rho_{ig} = \rho(D(G_i(\hat{\bm{z}}, \hat{\bm{y}})))$ and $\rho_{gi} = \rho(D_i(G(\hat{\bm{z}}, \hat{\bm{y}})))$.

Further, $\mathcal{L}_{kl, 1}$, $\mathcal{L}_{kl, 2}$ and $\mathcal{L}_{kl, 3}$ form the following overall distillation objective on the server side:
\begin{align}
    \label{KL_server:}
     & \min_{\bm{w}, \bm{\theta}_D} \mathbbm{E}_{\hat{\bm{z}}, \hat{y}\sim \mathcal{N}(\bm{0},\bm{I}), p(y)}[\mathcal{L}_{kl, 1} + \mathcal{L}_{kl, 2} + \mathcal{L}_{kl, 3}].
\end{align}

\subsection{Discussion}

\textbf{Privacy.} 
FedMD-CG trains a local generator on each client for replacing the local feature extractor~(LFE) by simulating the output vector space of LFE, i.e., the latent feature space, rather than the distribution space of private data. 
In other words, the local generator captures only high-level feature patterns of the local model, which are incomprehensible to human beings.
Also, FedMD-CG requires each client to share its classifier.
In our work, the classifier is in the top layers~(i.e., fully connected layers) with a high degree of abstraction.
As verified by~\cite{Yosinski2014How},  the lower layer features are more general and higher layer features have larger specificity.
This suggests that different inputs to the model can result in the same top-layer activations, making it difficult to reconstruct the original data with the classifier~\cite{Wang2021Data}.
Therefore, FedMD-CG can reduce the risk of privacy leakage, and has the same level of privacy protection as FedCG.

\textbf{Consistency and Computing cost.}
FedMD-CG performs client-side knowledge transfer at the latent feature level and the logit level, thus extracting knowledge embedded in the global generator 
and local model more directly and efficiently than FedCG.
Meanwhile, our method guarantees the consistency of the local generator and classifier trained by each client, which may not be satisfied in FedCG. 
To put it differently, FedMD-CG requires the local generator to generate pseudo-features that the local classifier can significantly distinguish in order to make the generator output more fidelity. 
According to our experiments, the consistency of FedMD-CG ensures high-quality aggregation on the server side and robustness to data heterogeneity. 
In addition, FedMD-CG does not employ an additional discriminator to adversarially train the local generator under the cGAN framework independently of the local classifier like FedCG, which reduces the client's computing cost. 

\section{Experiments}
\label{experiment:}

\subsection{Implementation Settings}
\label{Imple_Set:}

\textbf{Datasets.}
We perform our experiments on three public datasets EMNIST~\cite{Cohen2017EMNIST}, Fashion-MNIST~\cite{xiao2017fashion}~(FMNIST in short in this paper), and CIFAR-10~\cite{Krizhevsky2009Learning}. 
Following existing works~\cite{Zhang2022Fine, Acar2021Federated, Zhu2021Data, luo2023decentralized},
we use Dirichlet process $Dp(\omega)$ to strictly partition the training set of each dataset across clients.
Notably, a smaller $\omega$ corresponds to higher data heterogeneity. We set $\omega \in \{0.1, 1.0, 10.0\}$ in our experiments. 

\textbf{Backbone Architectures and Baselines.}
Throughout all our experiments, we deploy LeNet-5~\cite{LeCun1998Gradient} as the backbone network with two convolutional layers~(i.e., feature extractor) and three fully connected layers~(i.e., classifier).
Similarly, we employ three fully connected layers with BatchNorm as the generator for each client and adjust its output dimension to match that of the corresponding feature extractor. 
We select five FL methods most relevant to our work as baselines for comparison, including FedAvg~\cite{McMahan2017Communication}, FedPer~\cite{Arivazhagan2019Federated}, LG-FedAvg~\cite{Liang2020Think}, FedGen~\cite{Zhu2021Data} and FedCG~\cite{Wu2021Fedcg}. 
Moreover, we consider the baseline that trains a local model for each client, without any sharing. We call it Local Training~(LT for short). 
For fairness, FedGen shares clients' classifiers with the server. 
In particular, we treat the client's classifier 
whose output dimension is set to $1$ as the discriminator of cGAN in FedCG.

\begin{table}[t]
  \centering
  \caption{Test performance~(\%) comparison between FedMD-CG and baselines over different datasets. Note that $L.acc$ and $G.acc$ denote \textit{local test accuracy} and \textit{global test accuracy}, respectively.}
  \resizebox{1.0\columnwidth}{!}{
    \begin{tabular}{ccc|cc|cc|cc|cc|cc}
    \toprule
    \multirow{3}[4]{*}{Alg.s} & \multicolumn{4}{c|}{EMNIST}   & \multicolumn{4}{c|}{FMNIST}   & \multicolumn{4}{c}{CIFAR-10} \\
\cmidrule{2-13}          & \multicolumn{2}{c|}{$\omega=1.0$} & \multicolumn{2}{c|}{$\omega=0.1$} & \multicolumn{2}{c|}{$\omega=1.0$} & \multicolumn{2}{c|}{$\omega=0.1$} & \multicolumn{2}{c|}{$\omega=10.0$} & \multicolumn{2}{c}{$\omega=1.0$} \\
          & $L.acc$ & $G.acc$ & $L.acc$ & $G.acc$ & $L.acc$ & $G.acc$ & $L.acc$ & $G.acc$ & $L.acc$ & $G.acc$ & $L.acc$ & $G.acc$ \\
    \midrule
    FedAvg  & \textbf{96.29$\pm$0.06} & 96.83$\pm$0.07 & \textbf{85.65$\pm$2.33} & \textbf{95.06$\pm$0.46} & \textbf{80.99$\pm$0.81} & \textbf{84.77$\pm$0.30} & \textbf{59.29$\pm$3.19} & \textbf{78.91$\pm$2.12} & 51.68$\pm$0.53 & 54.18$\pm$0.52 & 42.84$\pm$2.03 & \textbf{52.39$\pm$0.58} \\
    LT    & 90.48$\pm$1.46 & 95.17$\pm$0.70 & 40.77$\pm$2.34 & 62.08$\pm$5.11 & 74.19$\pm$2.92 & 80.32$\pm$1.02 & 37.71$\pm$2.99 & 56.34$\pm$10.42 & 49.83$\pm$0.88 & 48.99$\pm$1.35 & 40.92$\pm$2.25 & 37.92$\pm$2.24 \\
    \midrule
    FedPer & 91.96$\pm$1.20 & 96.45$\pm$0.10 & 41.08$\pm$2.40 & 76.90$\pm$2.34 & 75.83$\pm$2.42 & 82.94$\pm$1.11 & 37.39$\pm$3.17 & 61.88$\pm$9.80 & 50.72$\pm$0.63 & 53.87$\pm$0.58 & 41.80$\pm$2.15 & 49.83$\pm$1.66 \\
    LG-FedAvg & 94.01$\pm$0.53 & 96.22$\pm$0.19 & 46.29$\pm$3.39 & 86.48$\pm$1.75 & 77.03$\pm$1.94 & 82.55$\pm$0.46 & 38.89$\pm$3.18 & 66.35$\pm$6.65 & 50.11$\pm$0.80 & 51.80$\pm$0.67 & 41.49$\pm$2.56 & 44.59$\pm$1.88 \\
    \midrule
    FedGen & 95.62$\pm$0.38 & 97.64$\pm$0.17 & 51.29$\pm$4.01 & 87.69$\pm$2.50 & 77.88$\pm$3.10 & 83.81$\pm$1.95 & 41.96$\pm$3.40 & 68.05$\pm$3.96 & 52.94$\pm$2.38 & 48.49$\pm$3.13 & 38.13$\pm$4.89 & 40.85$\pm$3.87 \\
    FedCG & 96.06$\pm$0.33 & \textbf{97.70$\pm$0.16} & 49.91$\pm$3.83 & 87.66$\pm$2.08 & 74.92$\pm$2.11 & 81.74$\pm$0.81 & 34.97$\pm$2.55 & 54.61$\pm$2.67 & 39.39$\pm$5.23 & 37.06$\pm$4.35 & 30.44$\pm$3.30 & 26.79$\pm$2.82 \\
    FedMD-CG & 95.45$\pm$0.25 & 97.18$\pm$0.17 & 54.45$\pm$3.56 & 87.87$\pm$1.64 & 79.00$\pm$1.43 & 84.47$\pm$0.38 & 42.55$\pm$3.68 & 71.09$\pm$1.01 & \textbf{54.82$\pm$0.79} & \textbf{55.18$\pm$1.75} & \textbf{46.30$\pm$2.24} & 47.56$\pm$2.21 \\
    \bottomrule
    \end{tabular}}%
  \label{table1:}%
\end{table}%

\begin{figure*}[h]\captionsetup[subcaption]{font=scriptsize}
	\centering
	\begin{subfigure}{0.329\linewidth}
		\centering
		\includegraphics[width=1.0\linewidth]{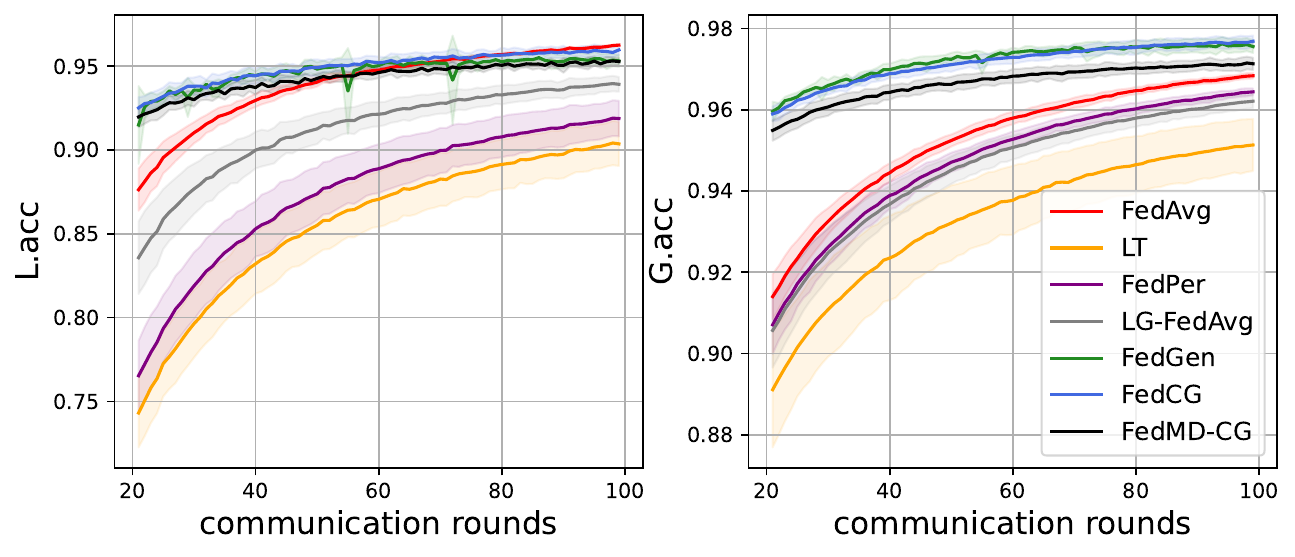}
		\caption{EMNIST, $\omega=1.0$}
		\label{fig3-a:}
	\end{subfigure}
        \centering
	\begin{subfigure}{0.329\linewidth}
		\centering
		\includegraphics[width=1.0\linewidth]{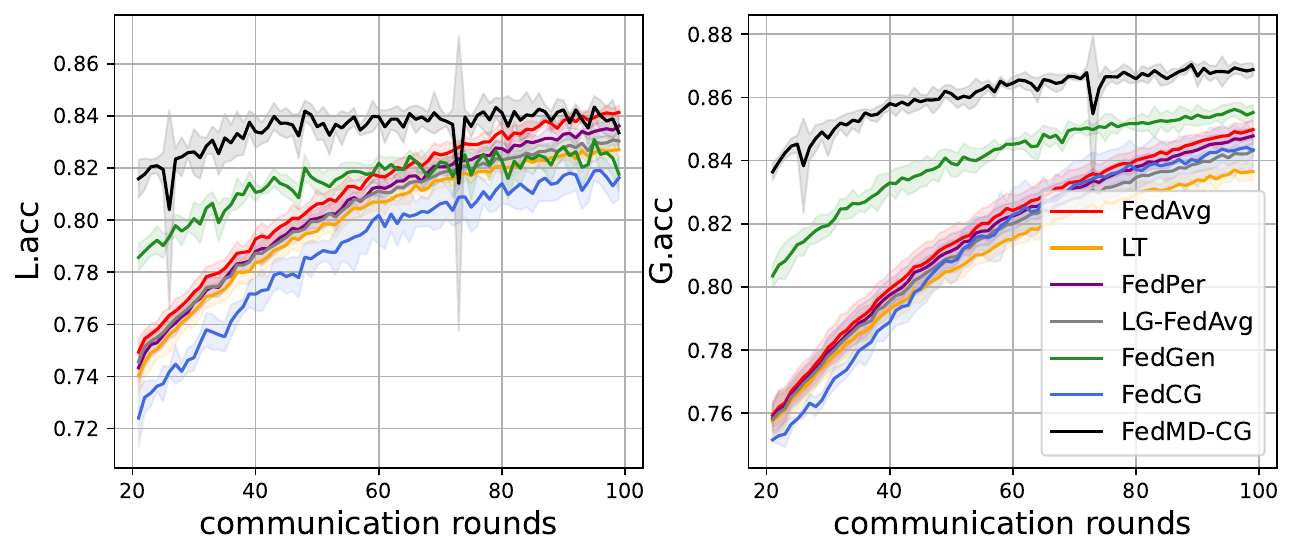}
		\caption{FMNIST, $\omega=10.0$}
		\label{fig3-a:}
	\end{subfigure}
	\centering
	\begin{subfigure}{0.329\linewidth}
		\centering
		\includegraphics[width=1.0\linewidth]{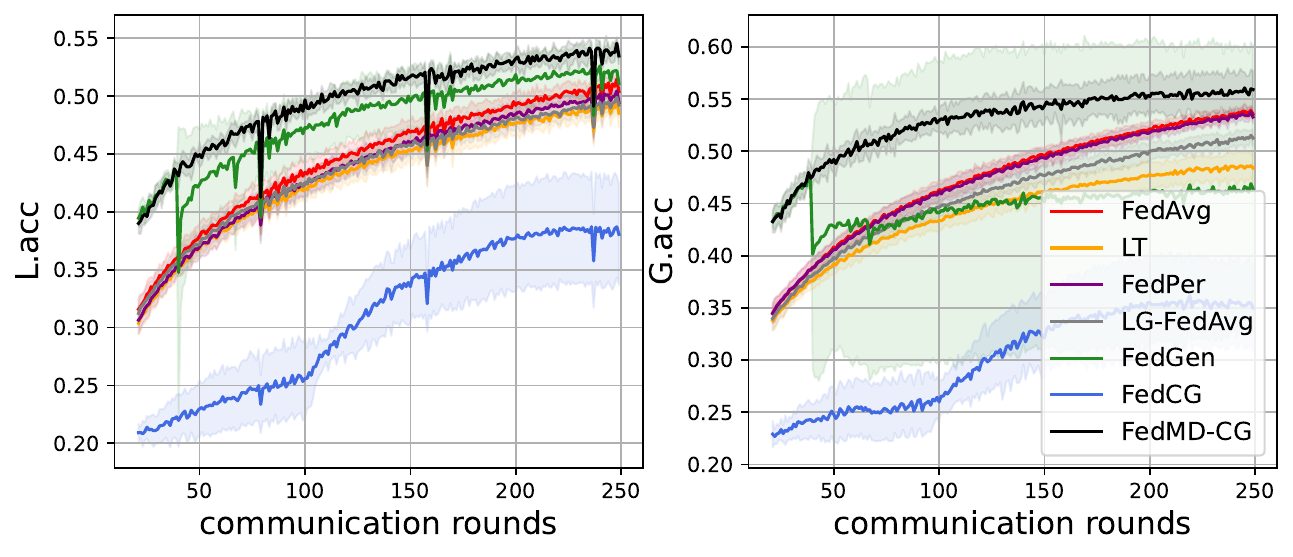}
		\caption{CIFAR-10, $\omega=10.0$}
		\label{fig3-c:}
	\end{subfigure} \\
        \centering
	\begin{subfigure}{0.329\linewidth}
		\centering
		\includegraphics[width=1.0\linewidth]{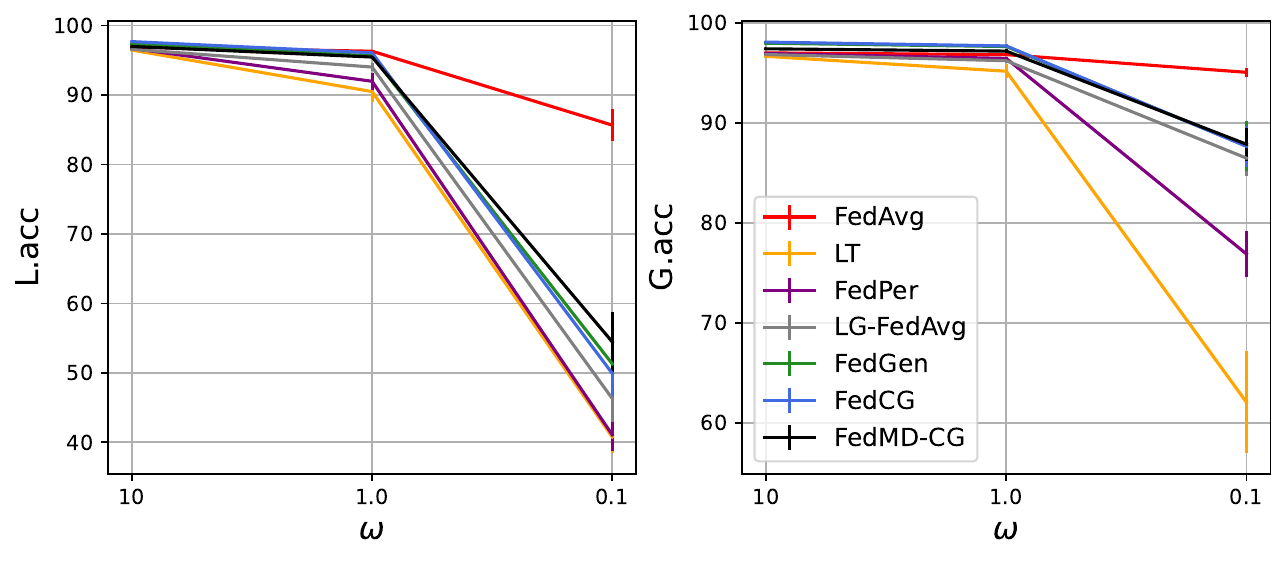}
		\caption{EMNIST}
		\label{fig3-a:}
	\end{subfigure}
	\centering
	\begin{subfigure}{0.329\linewidth}
		\centering
		\includegraphics[width=1.0\linewidth]{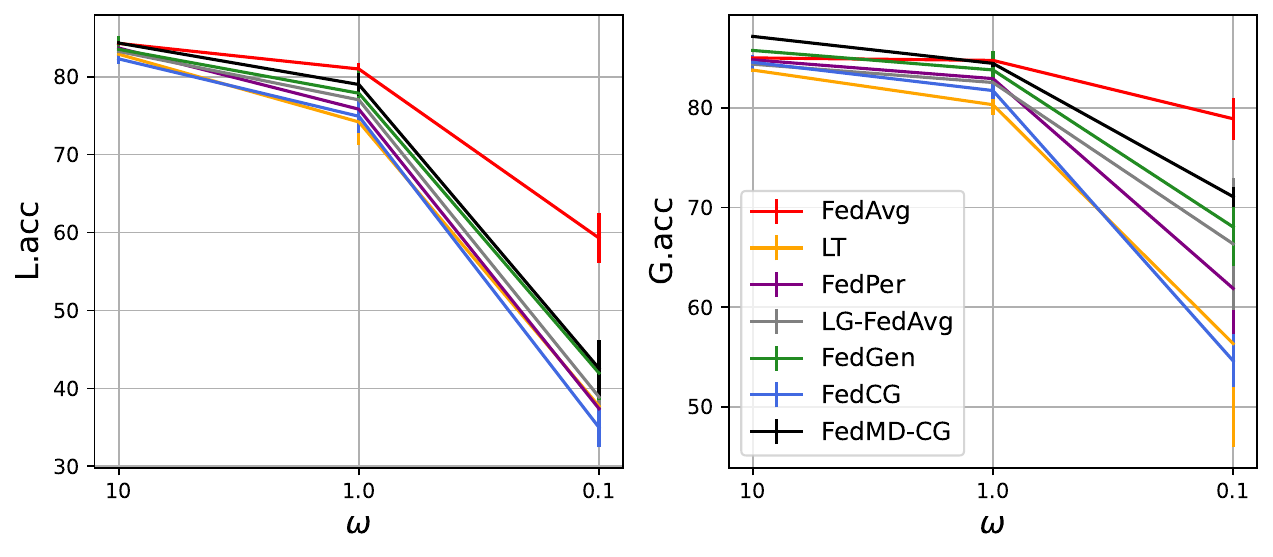}
		\caption{FMNIST}
		\label{fig3-b:}
	\end{subfigure}
        \centering
	\begin{subfigure}{0.329\linewidth}
		\centering
		\includegraphics[width=1.0\linewidth]{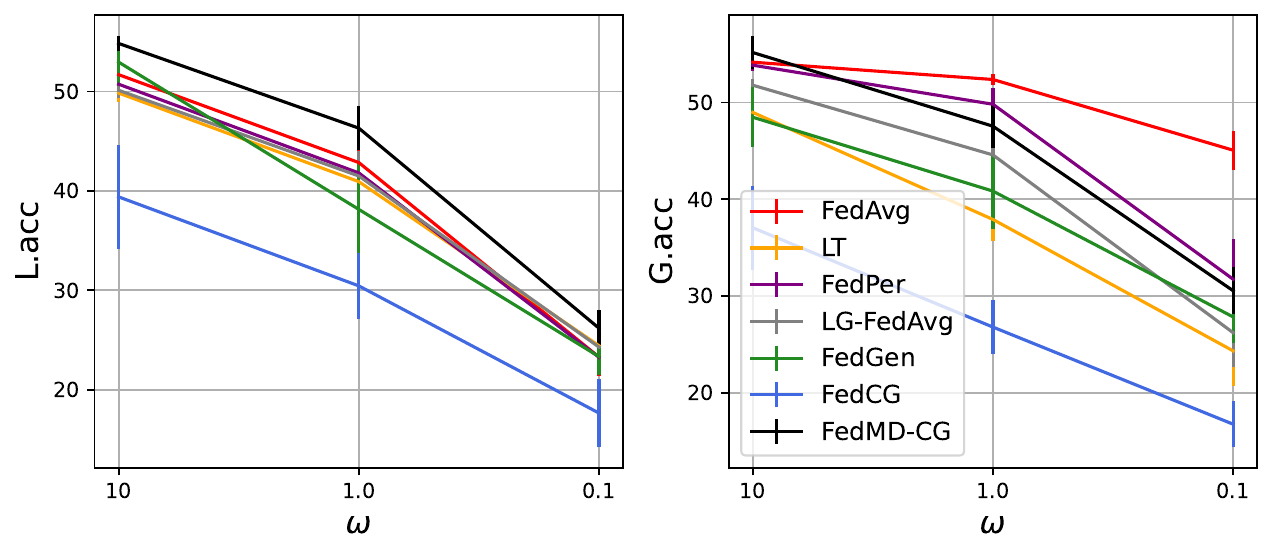}
		\caption{CIFAR-10}
		\label{fig3-c:}
	\end{subfigure}
	\caption{(a)-(c) are learning curves selected from FedMD-CG as well as baselines over different datasets. (d)-(f) show test performance~(\%)  w.r.t data hetergeneity over each dataset.} 
	\label{fig3:}
\end{figure*}


\textbf{Evaluation Metrics.}
We use the test set of each dataset to evaluate the test performance of different FL methods. 
1)~\textit{Local test accuracy.}~We randomly and evenly distribute the test set to each client and harness the test set on each client to verify the performance of local models.
2)~\textit{Global test accuracy}. 
We construct a virtual global model to evaluate the global performance of different FL methods via utilizing the original test set. As with FedAvg, this virtual global model is obtained by uploading all local models to the server for weighted average. 
3) {\it Peak signal-to-noise ratio~(PSNR)}.
Consistent with FedCG~\cite{Wu2021Fedcg}, we also consider the server is malicious, which uses DLG attack~\cite{Zhu2019Deep} to recover the original data from victim clients. We employ PSNR to measure the quality of the recovered images, thus 
evaluating the privacy-preserving capability of different FL methods.
To ensure reliability, we report the average for each experiment over $5$ different random seeds.
Due to the space limitations, we relegate full experimental settings and results to Appendix~\ref{app1:}.

\subsection{Results Comparison}
\label{Performance_Comp:}

\begin{wraptable}{r}{7.3cm}
  \centering
  \caption{Comparison of FedMD-CG and baselines in terms of PSNR~(dB)~($\omega=10.0$). Note that both FedGen and LG-FedAvg upload local classifiers to the server, and their privacy-preserving capabilities are intuitively the same, so we only report the PSNR of LG-FedAvg.}
    \resizebox{0.5\columnwidth}{!}{
    \begin{tabular}{c|c|c|c}
    \toprule
    \multirow{1}[2]{*}{Alg.s} & EMNIST & FMNIST & CIFAR-10 \\
    \midrule
    FedAvg & 24.54$\pm$0.15 & 22.63$\pm$0.61 & 29.22$\pm$1.56 \\
    FedPer & 23.55$\pm$0.52 & 19.66$\pm$1.03 & 14.84$\pm$2.61 \\
    LG-FedAvg & \textbf{6.78$\pm$0.09} & \textbf{6.33$\pm$1.32} & \textbf{8.75$\pm$1.03} \\
    FedCG & 7.05$\pm$0.63 & 6.98$\pm$1.57 & 9.87$\pm$1.83 \\
    FedMD-CG & 6.95$\pm$0.31 & 7.02$\pm$1.22 & 9.69$\pm$1.04 \\
    \bottomrule
    \end{tabular}}%
  \label{table2:}%
\end{wraptable}%
\textbf{Overview test performance comparison.} 
As shown in Table~\ref{table1:}, FedAvg achieves the best test performance while FedMD-CG achieves the second-best test performance in most cases of EMNIST and FMNIST.
FedAvg's test performance benefits from the fact that the server can collect complete local models from clients and then obtain the real global model to ensure remarkable test performance.
In most cases, the test performance of LT is worse than that of other methods since no information is shared among clients, inevitably causing over-fitting and poor generalization to new samples.
LG-FedAvg consistently outperforms FedPer w.r.t the local test accuracy, indicating that personalized classifiers can mitigate the sacrifice of local model performance when the feature extractor has several convolutional layers.
Meanwhile, FedMD-CG achieves the optimal local test accuracy on CIFAR-10.
We conjecture that the simple model average aggregation in FedAvg may counteract the personalized knowledge from clients,
thus adversely affecting local models' performance in difficult classification tasks.
Further, Fig.~\ref{fig3:}~(b)-(e) demonstrate that there is an overwhelming advantage of FedMD-CG over baselines in terms of learning efficiency during the early stages of training. Particularly, the local learning efficiency of FedMD-CG consistently outperforms that of baselines on FMNIST with $\omega=10.0$ and CIFAR-10.
Fig.~\ref{fig3:}~(f)-(h) reveal the impact of data heterogeneity on test performance for the methods. It can be observed that the test performance of all methods deteriorates as $\omega$ decreases.
In most cases, FedMD-CG dominates the baselines that share only part of the model in terms of the local test accuracy.
Also, FedMD-CG uniformly surpasses baselines w.r.t the local test accuracy over varying $\omega$ on CIFAR-10. 
This indicates that our method is robust to data heterogeneity.

\begin{wrapfigure}{r}{0cm}
\centering
\includegraphics[width=0.5\textwidth]{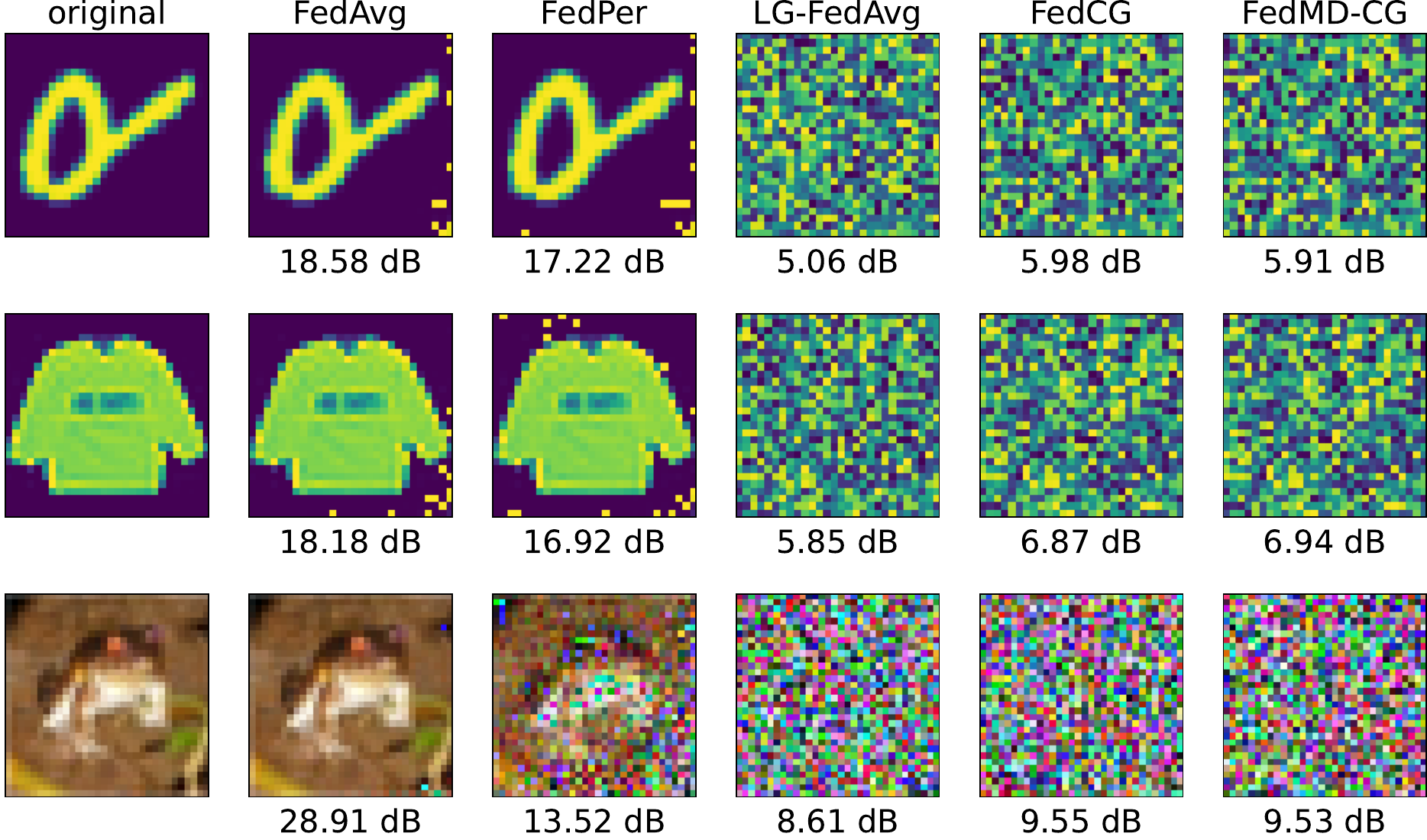} 
\caption{Image reconstruction with DLG attack in FedMD-CG and baselines. From the first to the last row, the images are selected from EMNIST, FMNIST and CIFAR-10 respectively. PSNR~(dB) is reported under each recovered image.}
\label{fig4:}
\vspace*{-3ex}
\end{wrapfigure}
\textbf{Privacy comparison.}
Here, we compare the privacy-preserving ability of FedMD-CG with other baselines under DLG attack.
It is worth noting that PSNR measures the similarity between the original image and the restored image. 
A larger PSNR value indicates a higher similarity between the images.
As observed in Table~\ref{table2:}, while FedAvg achieves excellent test performance~(see Table~\ref{table1:}), it scores the highest PSNR value across all datasets, which seriously threatens clients' private information.
Also, Fig.~\ref{fig4:} illustrates that the DLG attack is able to reconstruct the image very close to the original image in FedAvg.
According to Table~\ref{table1:} and Fig.~\ref{fig4:}, it is noticed that the strategy in FedPer to share clients' feature extractors should be prohibited, as it neither enables competitive test performance nor protects clients' privacy.
On the other hand, 
LG-FedAvg, FedCG and FedMD-CG can effectively prevent the privacy leakage of clients due to the low PSNR values.
Despite the small performance gap between FedMD-CG and LG-FedAvg w.r.t. PSNR,
FedMD-CG can significantly
outperform LG-FedAvg in terms of test performance (see Table~\ref{table1:}).
This indicates that approximating the local feature extractor with a generator not only has little privacy leakage risk but also improves performance.

\begin{wrapfigure}{r}{0cm}
\centering
\includegraphics[width=0.5\textwidth]{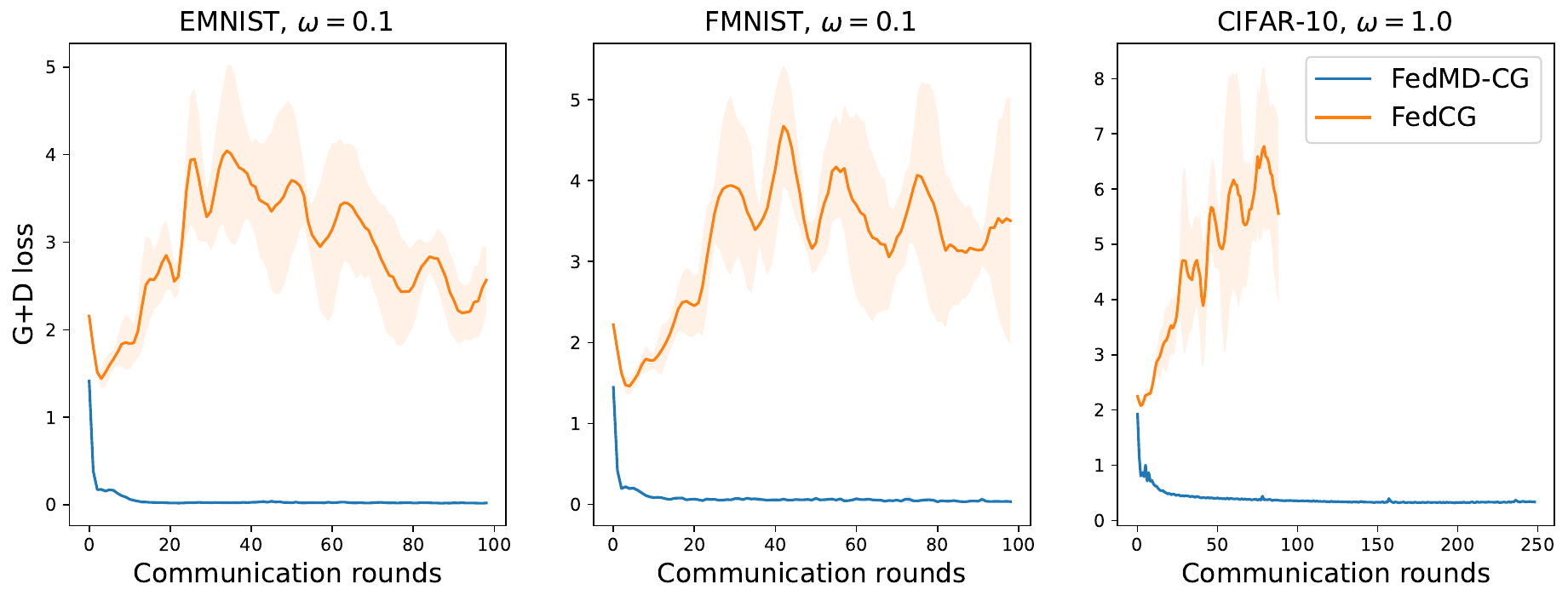} 
\caption{The consistency comparison between local generators and classifiers for FedCG and FedMD-CG w.r.t. \textit{AVE\_agg$^\star$}. G+D loss denotes the classification loss of the local classifier on the output of the local generator.}
\label{Fig5:}
\vspace*{-3ex}
\end{wrapfigure}
\textbf{Comparison between FedCG and FedMD-CG.}
From Table~\ref{table1:} and Fig.~\ref{fig3:}, the test performance of FedCG is worse than that of FedMD-CG in most cases, even worse than other baselines on CIFAR-10 and FMNIST ($\omega=0.1$).
We speculate that this attributes to the way FedCG transfers knowledge from the global generator to the local model and the inconsistency between the local generator and classifier in each client.
We next perform extensive experiments to verify our statement, as shown in Table~\ref{table3:} and Fig.~\ref{Fig5:}.
From Table~\ref{table3:}, FedMD-CG with \textit{AVE\_agg$^\star$} consistently surpasses FedCG with \textit{AVE\_agg$^\star$} in terms of the local test accuracy. 
The main reason is that FedMD-CG enables the local generators to extract the knowledge of the local models more effectively, which results in higher-quality trained local generators.
Also, the test performance of FedMD-CG with \textit{AVE\_agg} uniformly leads that of FedMD-CG with \textit{AVE\_agg$^\star$}, suggesting the insufficiency of knowledge transfer from the global generator to local models only at the latent feature level.
In addition, we compare the efficacy of different server-side aggregation manners.
Concretely, the test performance of FedMD-CG with \textit{KDC\_agg} consistently outperforms FedMD-CG with \textit{KD\_agg}, indicating that \textit{KDC\_agg} is more effective in transferring knowledge from local generators and classifiers to the global generator and classifier.
However, \textit{KD\_agg} and \textit{KDC\_agg} significantly deteriorate the test performance of FedCG.
We conjecture that the output of the local generator does not match the local classifier's, that is, the local classifier cannot effectively distinguish the output of the local generator, resulting in the degraded performance of FedCG. 
As shown in Fig.~\ref{Fig5:}, G+D loss of FedCG is consistently larger than that of FedMD-CG and does not converge. 
In general, the inconsistency of the local generator and classifier in each client may impede the server-side knowledge distillation aggregation training, resulting in poor performance of FedCG.

\begin{table}[t]
  \centering
  \caption{Test performance~(\%) comparison FedMD-CG and FedCG with different server-side aggregation manners over different datasets. Note that \textit{AVE\_agg} and \textit{AVE\_agg$^\star$} denote weighted average aggregation.
  Specifically, FedMD-CG with \textit{AVE\_agg} transfers the knowledge from the global generator to local models at both the latent feature level and the logit level, whereas FedMD-CG with \textit{AVE\_agg$^\star$} transfers the knowledge from the global generator to local models only at the latent feature level.
  Also, \textit{KD\_agg} and \textit{KDC\_agg} denote the server-side aggregation manners from FedCG and FedMD-CG, respectively.}
  \resizebox{1.0\columnwidth}{!}{
    \begin{tabular}{c|l|ll|ll|ll}
    \toprule
        \multirow{2}[1]{*}{Alg.s} & \multicolumn{1}{c|}{\multirow{2}[1]{*}{ Agg.}} & \multicolumn{2}{c|}{EMNIST, $\omega=0.1$} & \multicolumn{2}{c|}{FMNIST, $\omega=0.1$} & \multicolumn{2}{c}{CIFAR-10, $\omega=1.0$} \\
          &       & \multicolumn{1}{c}{$L.acc$} & \multicolumn{1}{c|}{$G.acc$} & \multicolumn{1}{c}{$L.acc$} & \multicolumn{1}{c|}{$G.acc$} & \multicolumn{1}{c}{$L.acc$} & \multicolumn{1}{c}{$G.acc$} \\
    \midrule
    \multirow{3}[2]{*}{FedCG} & \textit{AVE\_agg$^\star$}  & 50.55$\pm$4.32 & 86.74$\pm$1.44 & 39.46$\pm$3.40 & 67.41$\pm$3.59 & 41.85$\pm$2.48 & 44.98$\pm$1.79\\
          & \textit{KD\_agg} & 49.91$\pm$3.83 & 87.66$\pm$2.08 & 34.97$\pm$2.55 & 54.61$\pm$2.67 & 30.44$\pm$3.30 & 26.79$\pm$2.82 \\
          & \textit{KDC\_agg} & 39.65$\pm$4.67 & 82.32$\pm$4.66 & 37.23$\pm$2.54 & 62.89$\pm$7.01 & 28.87$\pm$0.90 & 25.92$\pm$1.53 \\
    \midrule
    \multirow{4}[2]{*}{FedMD-CG} & \textit{AVE\_agg$^\star$}  & 51.36$\pm$3.63 & 86.88$\pm$1.53 & 40.55$\pm$3.55 & 67.34$\pm$5.41 & 43.24$\pm$2.32 & 45.10$\pm$1.44 \\
          & \textit{AVE\_agg}  & 52.62$\pm$3.74 & 86.92$\pm$1.35 & 41.44$\pm$2.98 & 67.88$\pm$6.07 & 45.16$\pm$2.35 & 46.72$\pm$2.32 \\
          & \textit{KD\_agg} & 53.14$\pm$4.73 & 83.86$\pm$2.10 & 41.79$\pm$3.54 & 64.68$\pm$4.31 & 45.12$\pm$2.30 & 46.98$\pm$2.60 \\
          & \textit{KDC\_agg} & \textbf{54.45$\pm$3.56} & \textbf{87.87$\pm$1.64} & \textbf{42.55$\pm$3.68} & \textbf{71.09$\pm$1.01} & \textbf{46.30$\pm$2.24} & \textbf{47.56$\pm$2.21} \\
    \bottomrule
    \end{tabular}}%
  \label{table3:}%
  \vspace*{-3ex}
\end{table}%

\subsection{Ablation Study}
\label{Ablation_Study:}
\textbf{Necessity of losses in client-side for FedMD-CG.} We look into the test performance of FedMD-CG on CIFAR-10  with $\omega=10.0$ after discarding some losses in Eqs.~(\ref{gen_t_local:}) and~(\ref{local_t_gen:}), respectively, as shown in Table~\ref{loss-impact:}.
We can see that removing any loss leads to worse performance, i.e., lower local test accuracy and global test accuracy.
Also, their joint absence can cause further degradation of test performance. 
A trend in losses is observed that the absence of a single loss leads to a drop in test performance, while the removal of multiple losses enlarges the drop.
In addition, it should be noted that dropping multiple losses in the local generator update leads to more severe test performance degradation compared to the local model update. 
This shows that well-trained local generators can effectively boost the performance of our method, while under-trained local generators hinder the training of models.  
\begin{table}
  \centering
  \caption{Impact of each loss for client-side training over CIFAR-10 with $\omega=10.0$. Note that L.M.U and L.G.U denote the local model update and the local generator update, respectively. Also, we omit the subscript $i$ of each loss for client $i$.}
  \resizebox{0.8\columnwidth}{!}{
    \begin{tabular}{ccc|ccc}
    \toprule
    \multicolumn{6}{c}{FedMD-CG (baseline)} \\
    \midrule
    \multicolumn{3}{c}{$L.acc$} & \multicolumn{3}{c}{$G.acc$} \\
    \midrule
    \multicolumn{3}{c}{\textbf{54.82$\pm$0.79}} & \multicolumn{3}{c}{\textbf{55.18$\pm$1.75}} \\
    \midrule
    \midrule
    L.M.U & $L.acc$ & $G.acc$ & L.G.U & $L.acc$ & $G.acc$ \\
    \midrule
    $-{\mathop{\mathcal{L}}\limits ^{\to}}_{ce}$ & 51.53$\pm$1.04 & 52.52$\pm$1.37 & $-{\mathop{\mathcal{L}}\limits ^{\gets}}_{mse}$ & 52.73$\pm$1.15 & 53.19$\pm$1.72 \\
    $-{\mathop{\mathcal{L}}\limits ^{\to}}_{mse}$ & 53.07$\pm$0.97 & 53.73$\pm$1.99 & $-{\mathop{\mathcal{L}}\limits ^{\gets}}_{ce}$ & 53.89$\pm$0.89 & 52.88$\pm$2.09 \\
    $-{\mathop{\mathcal{L}}\limits ^{\to}}_{kl}$ & 53.46$\pm$0.94 & 53.34$\pm$1.74 & $-\mathcal{L}_{div}$ & 52.66$\pm$0.77 & 53.11$\pm$1.72 \\
    $-{\mathop{\mathcal{L}}\limits ^{\to}}_{ce}$, $-{\mathop{\mathcal{L}}\limits ^{\to}}_{mse}$ & 51.02$\pm$0.52 & 52.96$\pm$1.01 & $-{\mathop{\mathcal{L}}\limits ^{\gets}}_{mse}$, $-{\mathop{\mathcal{L}}\limits ^{\gets}}_{ce}$ & 46.94$\pm$1.33 & 49.37$\pm$1.53 \\
    $-{\mathop{\mathcal{L}}\limits ^{\to}}_{ce}$, $-{\mathop{\mathcal{L}}\limits ^{\to}}_{kl}$ & 51.55$\pm$0.60 & 52.81$\pm$1.22 & $-{\mathop{\mathcal{L}}\limits ^{\gets}}_{mse}$, $-{\mathcal{L}}_{div}$ & 47.80$\pm$0.30 & 50.15$\pm$1.24 \\
    $-{\mathop{\mathcal{L}}\limits ^{\to}}_{mse}$, $-{\mathop{\mathcal{L}}\limits ^{\to}}_{kl}$ & 52.64$\pm$0.40 & 53.46$\pm$1.25 & $-{\mathop{\mathcal{L}}\limits ^{\gets}}_{ce}$, $-{\mathcal{L}}_{div}$ & 48.02$\pm$0.31 & 50.41$\pm$1.61 \\
    $-{\mathop{\mathcal{L}}\limits ^{\to}}_{ce}$, $-{\mathop{\mathcal{L}}\limits ^{\to}}_{mse}$, $-{\mathop{\mathcal{L}}\limits ^{\to}}_{kl}$ & 50.27$\pm$0.41 & 49.55$\pm$1.28 & $-{\mathop{\mathcal{L}}\limits ^{\gets}}_{mse}$, $-{\mathop{\mathcal{L}}\limits ^{\gets}}_{ce}$, $-{\mathcal{L}}_{div}$ & 44.33$\pm$1.38 & 47.66$\pm$1.35 \\
    \bottomrule
    \end{tabular}}%
  \label{loss-impact:}%
  \vspace*{-3ex}
\end{table}%

\begin{wraptable}{r}{7.3cm}
  \centering
  \caption{Test performance~(\%) comparison among different diversity constraints. }
  \resizebox{0.5\columnwidth}{!}{
    \begin{tabular}{c|cc|cc|cc}
    \toprule
    \multirow{2}[2]{*}{Div. con.} & \multicolumn{2}{c|}{EMNIST, $\omega=0.1$} & \multicolumn{2}{c|}{FMNIST, $\omega=1.0$} & \multicolumn{2}{c}{CIFAR-10, $\omega=10.0$} \\
          & $L.acc$ & $G.acc$ & $L.acc$ & $G.acc$ & $L.acc$ & $G.acc$ \\
    \midrule
    $\mathcal{L}_{div}^0$  & 53.09$\pm$4.27 & \textbf{88.85$\pm$1.31} & 78.58$\pm$1.58 & 84.69$\pm$0.46 & 54.24$\pm$0.72 & 54.78$\pm$1.88 \\
    $\mathcal{L}_{div}^1$  & 53.65$\pm$4.12 & 88.51$\pm$0.80 & \textbf{79.03$\pm$1.52} & \textbf{84.73$\pm$0.49} & 54.81$\pm$0.71 & 54.90$\pm$1.73 \\
    $\mathcal{L}_{div}^2$  & \textbf{54.45$\pm$3.56} & 87.87$\pm$1.64 & 79.00$\pm$1.43 & 84.47$\pm$0.38 & \textbf{54.82$\pm$0.79} & \textbf{55.18$\pm$1.75} \\
    \bottomrule
    \end{tabular}}%
  \label{table4:}%
\end{wraptable}%
\textbf{Impacts of diversity constraints.} 
We also explore the effect of different diversity constraints on FedMD-CG.
Note that we omit the subscript $i$ of diversity loss for client $i$.
From Table~\ref{table4:}, FedMD-CG with $\mathcal{L}_{div}^1$ and $\mathcal{L}_{div}^2$ beats FedMD-CG with $\mathcal{L}_{div}^0$ w.r.t. the test performance in most case.
Also, $\mathcal{L}_{div}^1$ and $\mathcal{L}_{div}^2$ uniformly trump $\mathcal{L}_{div}^0$ in terms of the local test accuracy. 
Consequently, an empirical finding can be derived that imposing more weight on the inter-class output pair distance of the local generator boosts the local models’ performance.


\begin{wrapfigure}{r}{0cm}
\centering
\includegraphics[width=0.5\textwidth]{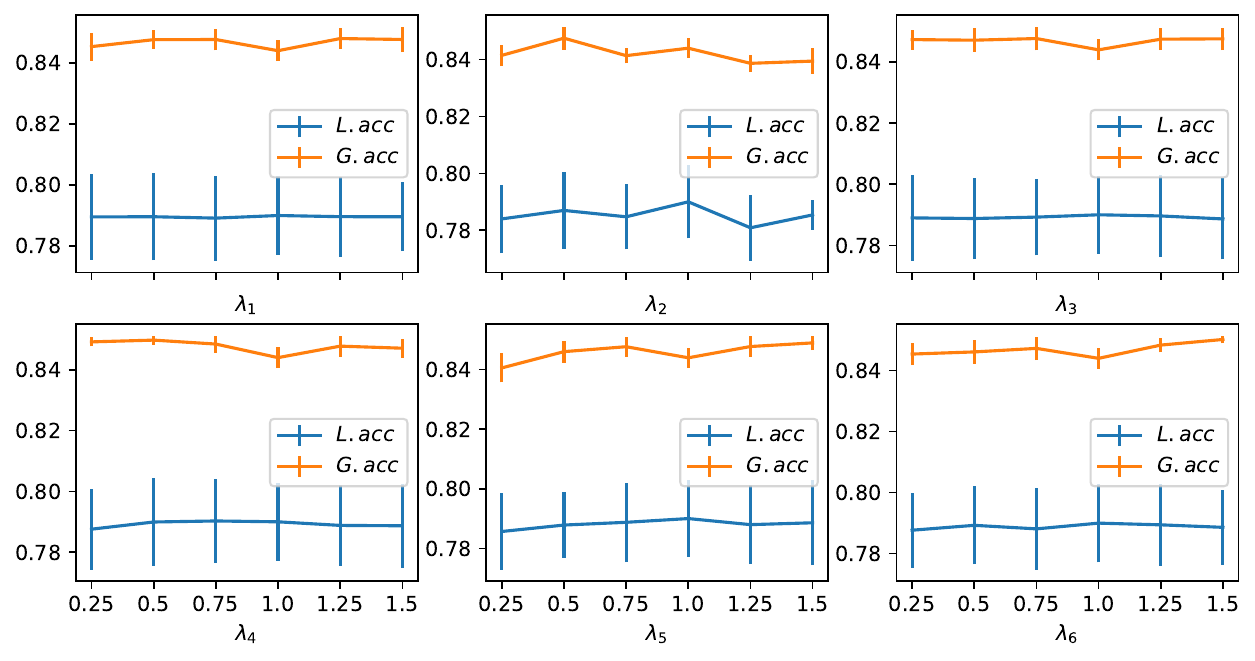} 
\caption{Test performance of FedMD-CG using varying hyperparameters on FMNIST with $\omega=1.0$.}
\label{fig6:}
\vspace*{-3ex}
\end{wrapfigure}
\textbf{Robustness of FedMD-CG against hyperparameters.} 
We investigate the test performance of FedMD-CG with varying hyperparameters over FMNIST.
We set $\omega=1.0$ and select 
$\lambda_{1}$, $\lambda_{2}$, $\lambda_{3}$, $\lambda_{4}$, $\lambda_{5}$ and $\lambda_{6}$ from $[0.25, 0.5, 0.75, 1.0, 1.25, 1.5]$. 
Fig.~\ref{fig6:} shows the test performance using the box plot, where FedMD-CG exemplifies similar test performance for non-zero selection of hyperparameters.
Notably, for a single loss, the effect of non-zero varying hyperparameters on the local test accuracy of FedMD-CG is slight.
This indicates that FedMD-CG is insensitive to the choice of non-zero hyperparameters over a large range for a single loss.

\section{Conclusions}
In this paper, we propose a novel FL method FedMD-CG, which achieves high competitive performance and high-level privacy preservation. 
Specifically, FedMD-CG decomposes each client's local model into a feature extractor and a classifier, and utilizes a conditional generator instead of the feature extractor to perform server-side model aggregation. 
Meanwhile, our method 
taps KD to train local models and generators at the latent feature level and the logit level, thereby ensuring the consistency of local generators and classifiers. 
Also, we construct additional classification losses and craft new diversity losses to enhance client-side training.
On the server side, FedMD-CG aggregates trained local generators and classifiers in a crossed data-free KD manner.
Finally, we conduct
extensive experiments to verify the superiority of FedMD-CG.
Due to space constraints, we discuss in detail the \textbf{limitations} and \textbf{broader impacts} of our work in Appendixes~\ref{app_discussion:} and~\ref{Broader_Impacts:}, respectively.

\clearpage
\appendix

\section*{Appendix}

\section{Related Work}
\label{Related Work:}

\textbf{Privacy Preservation in FL.} FL~\cite{McMahan2017Communication, Li2020Federated}
has emerged as a de facto machine learning area and received rapidly 
increasing research interest 
from the community. 
One of the primary attractions of FL is that it provides basic-level 
data privacy and security, as clients jointly train a global model by sharing model parameters or gradient updates without exposing their private data.
Yet, for such ``naked'' 
FL methods that do not provide any formal or provable privacy guarantees, some inference attacks (e.g., model-inversion~\cite{Yin2020Dreaming, Haim2022Reconstructing} and deep leakage from gradients~(DLG)~\cite{Zhu2019Deep, Geiping2020Inverting}) can easily extract sensitive information and even recover the original data from the trained model parameters or gradient updates without any information assistance. 
To reduce the risk of privacy leakage, FL often combines homomorphic encryption~(HE)~\cite{Gentry2009A} and differential privacy~(DP)~\cite{Dwork2006Calibrating}.
However, HE is computationally inefficient, while DP can deteriorate the training performance.
Meanwhile, there exists an alternative line 
of FL methods~\cite{Arivazhagan2019Federated, Liang2020Think, Shen2022Cd2}, which focuses 
on decomposing a model into private and public layers and sharing the public layers to boost the privacy of FL at the cost of inevitable performance drop. 
Particularly, FedPer~\cite{Arivazhagan2019Federated} splits a model into base and top layers. 
Each client uploads the base layers and hides the top layers from the server. Whereas, LG-FedAvg~\cite{Liang2020Think} shares the top layers while keeping the base layers localized.   

\textbf{Knowledge Distillation in FL.} 
The main insight of KD is to extract knowledge from one or more teacher models to a student model via learning their soft predictions, attention maps or intermediate~(latent) features~\cite{Hinton2015Distilling, Zagoruyko2016Paying, Yim2017A}. 
FL with KD has recently emerged as effective methods for dealing with real-world tasks.
For example, FedMLB~\cite{Kim2022Multi} and RHFL~\cite{Fang2022Robust} mitigate the fall 
of performance caused by data heterogeneity (i.e., non-IID). 
FedMD~\cite{Li2019Fedmd},
FedDP~\cite{Lin2020Ensemble} and
FCCL~\cite{Huang2022Learn} are able to perform FL with heterogeneous local models across clients.
In addition, FedGen\footnote{In this paper, we consider FedGen with partial parameter sharing.}~\cite{Zhu2021Data}
and FedKD~\cite{Gong2022Preserving} facilitate privacy-preserving of FL.

\textbf{Conditional Generator in FL.}
The objects mimicked by a conditional generator 
in FL can be roughly divided into two categories: clients' raw data and models' latent features.
For \textbf{the former}, FAug
~\cite{Jeong2018Communication} requires each client to collectively train a cGAN to augment its local data yielding an IID dataset.
DENSE~\cite{Zhang2022DENSE} and FedFTG~\cite{Zhang2022Fine} employ data-free KD to train a conditional generator on the server, thus training and fine-tuning the global model, respectively. 
Note that data-free KD is a promising approach to transfer knowledge from the teacher model 
to another student model without any real data~\cite{Chen2019Data, Fang2019Data}.
The mentioned FL methods utilize a conditional generator 
to enhance the generalization performance of global models under heterogeneity or communication cost constraints, but they are highly susceptible to privacy attacks or even violate the key privacy assumptions of FL.
Alternatively, the conditional generator 
can also be used as an attack tool
to reconstruct the private data of the victim clients on a malicious server~\cite{Li2022Auditing}.
For \textbf{the latter}, FedGen~\cite{Zhu2021Data} uploads the last layer of the local models to the server and trains a global conditional generator using data-free KD to boost the local model update of each client. FedCG~\cite{Wu2021Fedcg} integrates cGAN into FL aiming to harness a conditional generator to replace the local feature extractor and upload it to the server together with the local classifier, thus maintaining high-level privacy protection.

\section{Pseudocode}
\label{sec:pseudo}
In this section, we detail the pseudocode
of FedMD-CG in Algorithm~\ref{alg:1}.

\begin{algorithm}[!t]
  \caption{FedMD-CG} 
  \label{alg:1}
\begin{algorithmic}[1]
  \State {\bfseries Input:} 
  communication round $R$, client number $N$, label distribution $p(y)$, client-side training step $I_c$, client-side learning rates $\eta_c^l$, $\eta_c^\omega$, client-side label counter $\{c_i\}_{i\in [N]}$, server-side training step $I_s$, server-side learning rate $\eta_s$, batch size $B$,  hyperparameters $\lambda_{1}$, $\lambda_{2}$, $\lambda_{3}$, $\lambda_{4}$, $\lambda_{5}$, $\lambda_{6}$. 
  \State Initialize $\bm{w}^i$ and $\bm{\theta}^i=[\bm{\theta}_F^i, \bm{\theta}_D^i]$ on the client $i$.
  \State Initialize $[\bm{w}, \bm{\theta}_D]$ on the server. 
  \For{$r=1,\cdots, R$}
        \State Server broadcasts ($\bm{w}$, $\bm{\theta}_D$) and $p(y)$ to the clients.
        \State \textbf{On clients:}
        \For{$i \in [N]$ parallel}
            \State $\bm{\theta}_D^i = \bm{\theta}_D$
            \For{$\tau = 1,\cdots, I_c$}
                \State Sample $\{\bm{x}_b, \hat{\bm{z}}_b, y_b\}_{b=1}^{B}$ and resample $\{\hat{\bm{z}}_b, \hat{y}_b\}_{b=1}^{B}$, where $\{\bm{x}_b, y_b\}\sim \{\bm{X}_i, \bm{Y}_i\}$, $\hat{\bm{z}}_b \sim \mathcal{N}(\bm{0}, \bm{I})$ and $\hat{y}_b \sim p(y)$.
                \State Update label counter $c_i$.
                \State Update $\bm{\theta}^i$ with $\eta_c^l$ according to Eq.~(\ref{gen_t_local:}).
            \EndFor
            \State \textbf{end for}
            \For{$\tau = 1,\cdots, I_c$}
                \State Sample $\{\bm{x}_b, \hat{\bm{z}}_b, y_b\}_{b=1}^{B}$, where $\{\bm{x}_b, y_b\}\sim \{\bm{X}_i, \bm{Y}_i\}$ and $\hat{\bm{z}}_b \sim \mathcal{N}(\bm{0}, \bm{I})$.
                \State Update $\bm{w}^i$ with $\eta_c^{\omega}$ according to Eq.~(\ref{local_t_gen:}).
            \EndFor
            \State \textbf{end for}
            \State Upload $[\bm{w}^i, \bm{\theta}_D^i]$ and $c_i$ to the server.
        \EndFor
        \State \textbf{On server:}
        \State $[\bm{w}, \bm{\theta}_D] \leftarrow\sum_{i\in[N]}\frac{n_i}{\sum_{j\in[N]}n_j} [\bm{w}^i, \bm{\theta}_D^i]$ and update $p(y)$ based on $\{c_i\}_{i\in[N]}$.
        \For{$\tau = 1,\cdots, I_s$}
            \State Sample $\{\hat{\bm{z}}_b, \hat{y}_b\}_{b=1}^{B}$, where $\hat{\bm{z}}_b \sim \mathcal{N}(\bm{0}, \bm{I})$ and $\hat{y}_b \sim p(y)$.
            \State Update $[\bm{w}, \bm{\theta}_D]$ with $\eta_s$ according to Eq.~(\ref{KL_server:}).
        \EndFor
        \State \textbf{end for}
  \EndFor
  \State \textbf{end for}
\end{algorithmic}
\end{algorithm}

\section{Algorithm Description}

Algorithm~\ref{alg:1} summarizes the training procedure of FedMD-CG.
Concretely, starting from the local model update, clients 
first sample two mini-batch data $\{\bm{x}_b, \hat{\bm{z}}_b, y_b\}_{b=1}^{B}$ and $\{\hat{\bm{z}}_b, \hat{y}_b\}_{b=1}^{B}$ to perform the local model update~(lines 9-13), and then train the local generator with re-sampled mini-batch data $\{\bm{x}_b, \hat{\bm{z}}_b, y_b\}_{b=1}^{B}$~(lines 14-17).
The trained local generators and classifiers are sent to the server.
The server subsequently aggregates these generators and classifiers by simple model averaging to form the preliminary global generator and classifier, and then trains the global generator and classifier by using sampled data $\{\hat{\bm{z}}_b, \hat{y}_b\}_{b=1}^{B}$~(lines 21-24).
Notably, the global generator is under-trained at the early stages of training, which may mislead the local model training.
Therefore, during the training phase, $\lambda_{3}$, $\lambda_{2}$ and $\lambda_{1}$ are first initialized to $0$ and increase to pre-defined values with the increase of communication round. Readers are referred to the experimental section for more details.

\section{Computing devices and platforms}
\begin{itemize}
    \item OS: Ubuntu 18.04.3 LTS
    \item CPU: Intel(R) Xeon(R) Gold 6126 CPU @ 2.60GHz
    \item CPU Memory: 256 GB.
    \item GPU: NVIDIA Tesla V100 PCIe
    \item GPU Memory: 32GB
    \item Programming platform: Python 3.7.4
    \item Deep learning platform: PyTorch 1.9.0
\end{itemize}

\section{Full Experiments}
\label{app1:}
\subsection{Full Experimental Setting}
\textbf{Datasets.}
We perform our experiments on three public datasets EMNIST~\cite{Cohen2017EMNIST}, Fashion-MNIST~\cite{xiao2017fashion}~(FMNIST in short in this paper), and CIFAR-10~\cite{Krizhevsky2009Learning}. 
Following existing works~\cite{Zhang2022Fine, Acar2021Federated, Zhu2021Data},
we use Dirichlet process $Dp(\omega)$ to strictly partition the training set of each dataset across clients.
Notably, a smaller $\omega$ corresponds to higher data heterogeneity. We set $\omega \in \{0.1, 1.0, 10.0\}$ in our experiments. 

\textbf{Backbone Architectures and Baselines.}
Throughout all our experiments, we deploy LeNet-5~\cite{LeCun1998Gradient} as the backbone network with two convolutional layers~(i.e., feature extractor) and three fully connected layers~(i.e., classifier).
Similarly, we employ three fully connected layers with BatchNorm as the generator for each client and adjust its output dimension to match that of the corresponding feature extractor. 
We select five FL methods most relevant to our work as baselines for comparison, including FedAvg~\cite{McMahan2017Communication}, FedPer~\cite{Arivazhagan2019Federated}, LG-FedAvg~\cite{Liang2020Think}, FedGen~\cite{Zhu2021Data} and FedCG~\cite{Wu2021Fedcg}. 
Moreover, we consider the baseline that trains a local model for each client, without any sharing. We call it Local Training~(LT for short). 
For fairness, FedGen shares clients' classifiers with the server. 
In particular, we treat the client's classifier 
whose output dimension is set to $1$ as the discriminator of cGAN in FedCG.

\textbf{Configurations.}
For EMNIST~(FMNIST), we set communication round $R=100~(100)$ and client number $N=20~(10)$. And we set $R=250$ and $N=10$ for CIFAR-10.
We adopt client-side training step $I_c=20$ and server-side training step $I_s=50$. 
For client-side training, SGD and Adam are applied to optimize the local models and generators, respectively.
The learning rate $\eta_c^l$ for SGD is searched over the range of $\{0.01, 0.05, 0.08\}$ and the best one is picked. And we set $\eta_c^\omega=0.0003$ for Adam. 
For server-side training, the Adam optimizer with $\eta_s=0.0003$ is used to update the global generator and classifier.
For all update steps, we set batch size $B$ to $64$ and weight decay to $1e-4$. For FedMD-CG, we set the diversity constraint 
to $\mathcal{L}_{div}^2$ unless otherwise specified.
For the hyperparameters used to balance different loss items, all are set to $1$ unless otherwise specified. 
Particularly, in the local model update, we initialize $\lambda_{3}$, $\lambda_{2}$ and $\lambda_{1}$ to $0$ and increase their values to pre-defined values with the increase of communication round to avoid the misleading caused by the under-trained global generator.  
We set the parameter values to be incremented by $\lambda= \lambda^{pre}((r-1)/R)^d$, where $\lambda^{pre}$ is a pre-defined value and $d$  controls how fast the parameter increases. We set $d=1$. 
The dimension of $\hat{\bm{z}}\sim \mathcal{N}(\bm{0}, \bm{I})$ is $128$ for all datasets.

\subsection{Full Experimental Results}

\begin{table*}[h]
  \centering
  \caption{Test performance~(\%) comparison between FedMD-CG and baselines over EMNIST. Note that $L.acc$ and $G.acc$ denote \textit{local test accuracy} and \textit{global test accuracy}, respectively.}
    \resizebox{1.0\columnwidth}{!}{
    \begin{tabular}{ccc|cc|cc}
    \toprule
    \multirow{3}[6]{*}{Alg.s} & \multicolumn{6}{c}{EMNIST} \\
\cmidrule{2-7}          & \multicolumn{2}{c|}{$\omega=10.0$} & \multicolumn{2}{c|}{$\omega=1.0$} & \multicolumn{2}{c}{$\omega=0.1$} \\
\cmidrule{2-7}          & $L.acc$ & $G.acc$ & $L.acc$ & $G.acc$ & $L.acc$ & $G.acc$ \\
    \midrule
    FedAvg & 96.88$\pm$0.08 & 97.00$\pm$0.08 & \textbf{96.29$\pm$0.06} & 96.83$\pm$0.07 & \textbf{85.65$\pm$2.33} & \textbf{95.06$\pm$0.46} \\
    LT    & 96.45$\pm$0.08 & 96.66$\pm$0.13 & 90.48$\pm$1.46 & 95.17$\pm$0.70 & 40.77$\pm$2.34 & 62.08$\pm$5.11 \\
    \midrule
    FedPer & 96.69$\pm$0.10 & 96.94$\pm$0.08 & 91.96$\pm$1.20 & 96.45$\pm$0.10 & 41.08$\pm$2.40 & 76.90$\pm$2.34 \\
    LG-FedAvg & 96.57$\pm$0.11 & 96.84$\pm$0.10 & 94.01$\pm$0.53 & 96.22$\pm$0.19 & 46.29$\pm$3.39 & 86.48$\pm$1.75 \\
    \midrule
    FedGen & 97.34$\pm$0.16 & 97.97$\pm$0.09 & 95.62$\pm$0.38 & 97.64$\pm$0.17 & 51.29$\pm$4.01 & 87.69$\pm$2.50 \\
    FedCG & \textbf{97.67$\pm$0.03} & \textbf{98.08$\pm$0.07} & 96.06$\pm$0.33 & \textbf{97.70$\pm$0.16} & 49.91$\pm$3.83 & 87.66$\pm$2.08 \\
    FedMD-CG & 96.97$\pm$0.05 & 97.41$\pm$0.11 & 95.45$\pm$0.25 & 97.18$\pm$0.17 & 54.45$\pm$3.56 & 87.87$\pm$1.64 \\
    \bottomrule
    \end{tabular}}%
  \label{tab:addlabel}%
\end{table*}%

\begin{figure*}[h]\captionsetup[subcaption]{font=scriptsize}
	\centering
	\begin{subfigure}{0.45\linewidth}
		\centering
		\includegraphics[width=1.0\linewidth]{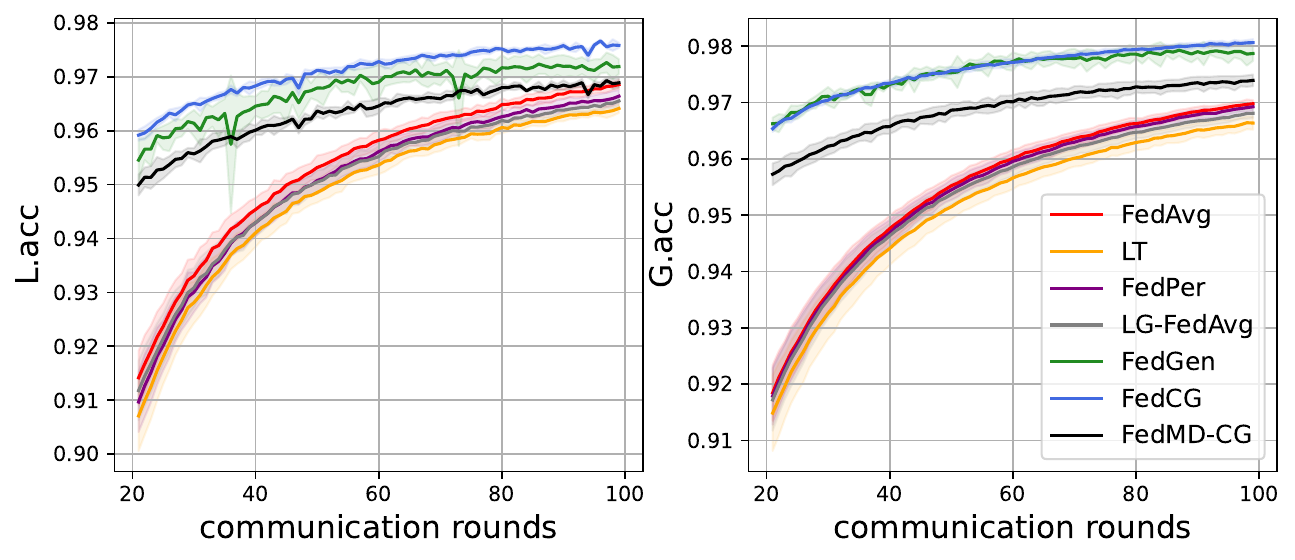}
		\caption{EMNIST, $\omega=10.0$}
		\label{fig3-a:}
	\end{subfigure}
	\centering
	\begin{subfigure}{0.45\linewidth}
		\centering
		\includegraphics[width=1.0\linewidth]{EMNIST_1.0_comm_round.pdf}
		\caption{EMNIST, $\omega=1.0$}
		\label{fig3-b:}
	\end{subfigure}
        \centering
	\begin{subfigure}{0.45\linewidth}
		\centering
		\includegraphics[width=1.0\linewidth]{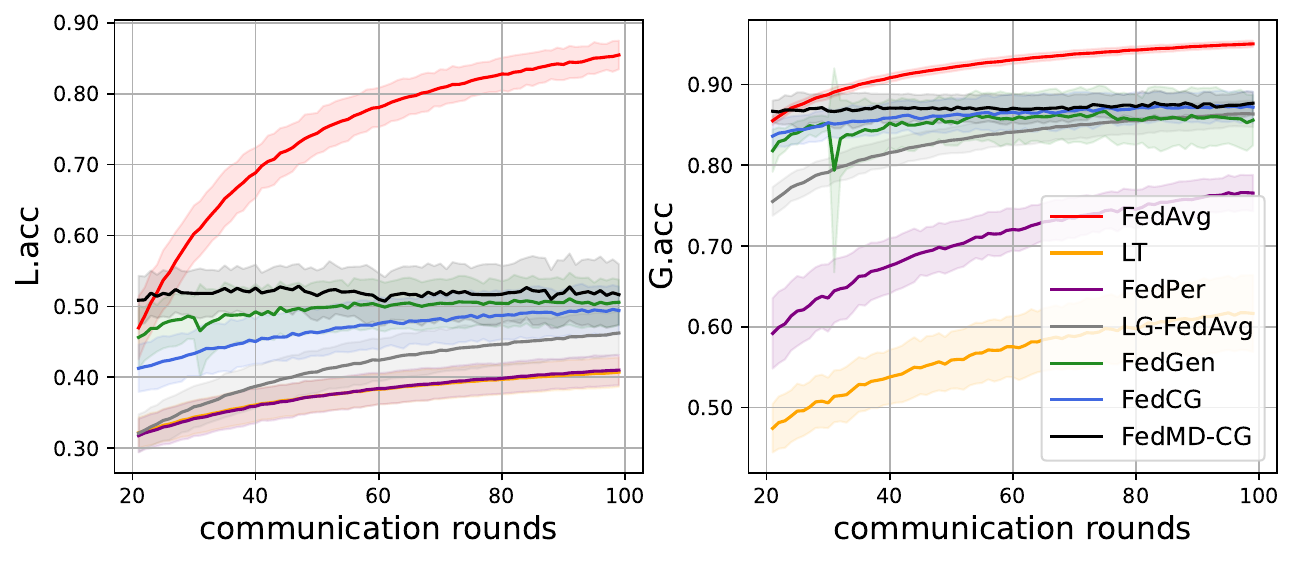}
		\caption{EMNIST, $\omega=0.1$}
		\label{fig3-c:}
	\end{subfigure}
	\caption{Learning curves for FedMD-CG as well as baselines over EMNIST. } 
        \vspace*{-2ex}
\end{figure*}

\begin{table}[h]
  \centering
  \caption{Test performance~(\%) comparison between FedMD-CG and baselines over FMNIST. Note that $L.acc$ and $G.acc$ denote \textit{local test accuracy} and \textit{global test accuracy}, respectively.}
    \resizebox{1.0\columnwidth}{!}{
    \begin{tabular}{ccc|cc|cc}
    \toprule
    \multirow{3}[6]{*}{Alg.s} & \multicolumn{6}{c}{FMNIST} \\
\cmidrule{2-7}          & \multicolumn{2}{c|}{$\omega=10.0$} & \multicolumn{2}{c|}{$\omega=1.0$} & \multicolumn{2}{c}{$\omega=0.1$} \\
\cmidrule{2-7}          & $L.acc$ & $G.acc$ & $L.acc$ & $G.acc$ & $L.acc$ & $G.acc$ \\
    \midrule
    FedAvg & 84.28$\pm$0.24 & 85.02$\pm$0.29 & \textbf{80.99$\pm$0.81} & \textbf{84.77$\pm$0.30} & \textbf{59.29$\pm$3.19} & \textbf{78.91$\pm$2.12} \\
    LT    & 82.89$\pm$0.29 & 83.80$\pm$0.23 & 74.19$\pm$2.92 & 80.32$\pm$1.02 & 37.71$\pm$2.99 & 56.34$\pm$10.42 \\
    \midrule
    FedPer & 83.68$\pm$0.21 & 84.83$\pm$0.27 & 75.83$\pm$2.42 & 82.94$\pm$1.11 & 37.39$\pm$3.17 & 61.88$\pm$9.80 \\
    LG-FedAvg & 83.25$\pm$0.24 & 84.39$\pm$0.19 & 77.03$\pm$1.94 & 82.55$\pm$0.46 & 38.89$\pm$3.18 & 66.35$\pm$6.65 \\
    \midrule
    FedGen & 83.52$\pm$1.76 & 85.77$\pm$0.12 & 77.88$\pm$3.10 & 83.81$\pm$1.95 & 41.96$\pm$3.40 & 68.05$\pm$3.96 \\
    FedCG & 82.29$\pm$0.63 & 84.58$\pm$0.77 & 74.92$\pm$2.11 & 81.74$\pm$0.81 & 34.97$\pm$2.55 & 54.61$\pm$2.67 \\
    FedMD-CG & \textbf{84.32$\pm$0.28} & \textbf{87.18$\pm$0.08} & 79.00$\pm$1.43 & 84.47$\pm$0.38 & 42.55$\pm$3.68 & 71.09$\pm$1.01 \\
    \bottomrule
    \end{tabular}}%
  \label{tab:addlabel}%
\end{table}%

\begin{figure*}[h]\captionsetup[subcaption]{font=scriptsize}
	\centering
	\begin{subfigure}{0.45\linewidth}
		\centering
		\includegraphics[width=1.0\linewidth]{FMNIST_10_comm_round.pdf}
		\caption{FMNIST, $\omega=10.0$}
		\label{fig3-a:}
	\end{subfigure}
	\centering
	\begin{subfigure}{0.45\linewidth}
		\centering
		\includegraphics[width=1.0\linewidth]{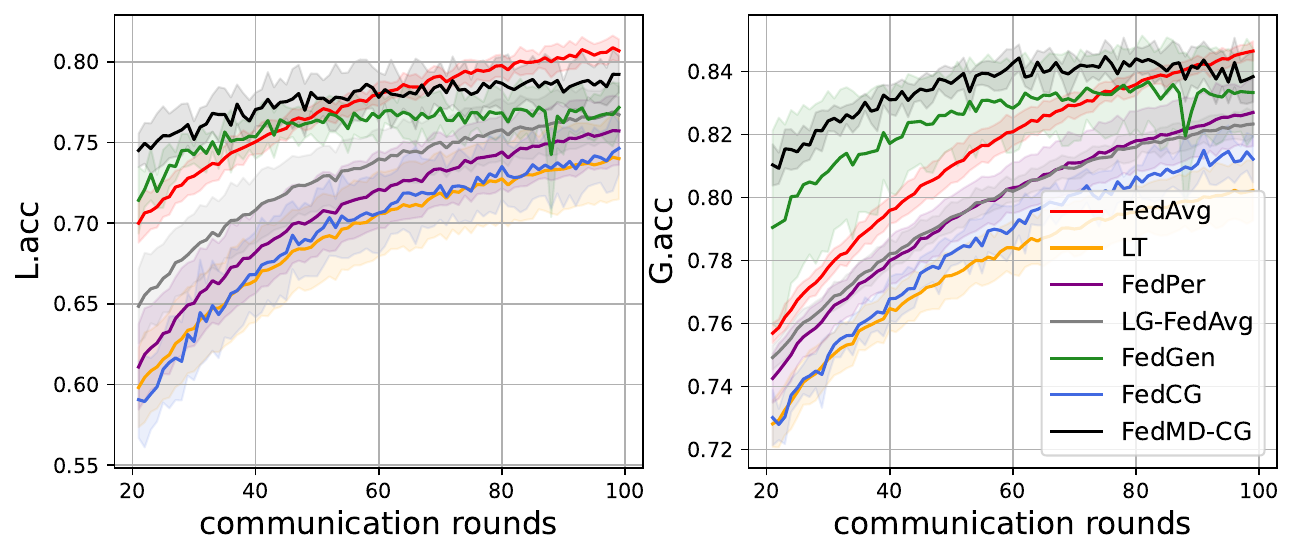}
		\caption{FMNIST, $\omega=1.0$}
		\label{fig3-b:}
	\end{subfigure}
        \centering
	\begin{subfigure}{0.45\linewidth}
		\centering
		\includegraphics[width=1.0\linewidth]{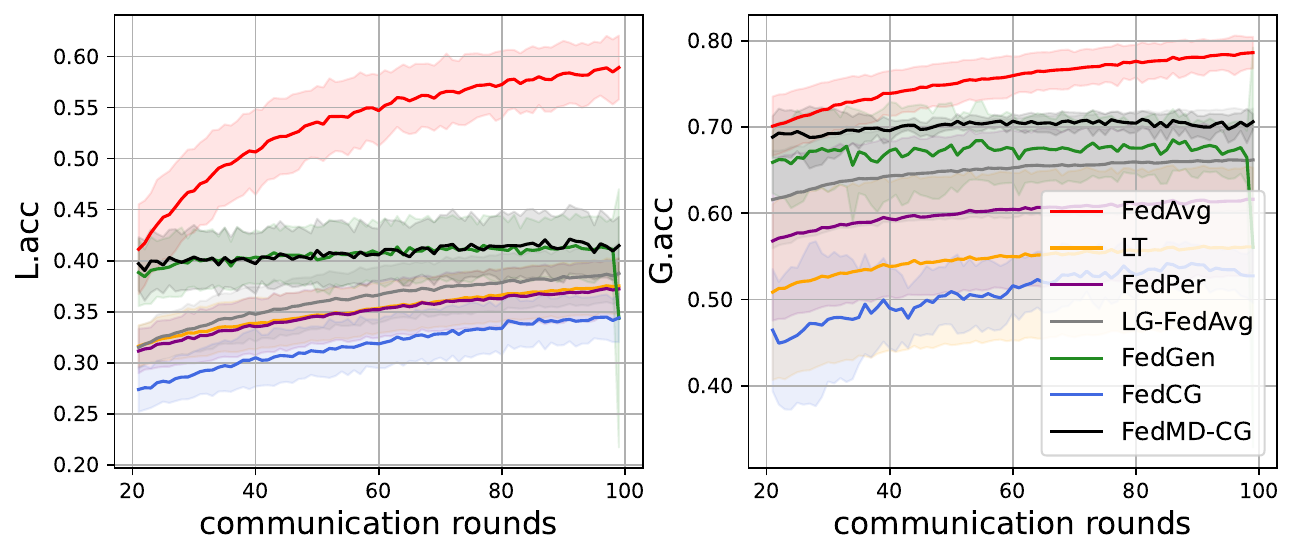}
		\caption{FMNIST, $\omega=0.1$}
		\label{fig3-c:}
	\end{subfigure}
	\caption{Learning curves for FedMD-CG as well as baselines over over FMNIST. } 
\end{figure*}

\begin{table*}[h]
  \centering
  \caption{Test performance~(\%) comparison between FedMD-CG and baselines over CIFAR-10. Note that $L.acc$ and $G.acc$ denote \textit{local test accuracy} and \textit{global test accuracy}, respectively.}
  \resizebox{1.0\columnwidth}{!}{
    \begin{tabular}{ccc|cc|cc}
    \toprule
    \multirow{3}[6]{*}{Alg.s} & \multicolumn{6}{c}{CIFAR-10} \\
\cmidrule{2-7}          & \multicolumn{2}{c|}{$\omega=10.0$} & \multicolumn{2}{c|}{$\omega=1.0$} & \multicolumn{2}{c}{$\omega=0.1$} \\
\cmidrule{2-7}          & $L.acc$ & $G.acc$ & $L.acc$ & $G.acc$ & $L.acc$ & $G.acc$ \\
    \midrule
    FedAvg & \multicolumn{1}{c}{51.68$\pm$0.53} & \multicolumn{1}{c|}{54.18$\pm$0.52} & \multicolumn{1}{c}{42.84$\pm$2.03} & \multicolumn{1}{c|}{\textbf{52.39$\pm$0.58}} & \multicolumn{1}{c}{23.28$\pm$1.94} & \multicolumn{1}{c}{\textbf{45.08$\pm$2.03}} \\
    LT    & \multicolumn{1}{c}{49.83$\pm$0.88} & \multicolumn{1}{c|}{48.99$\pm$1.35} & \multicolumn{1}{c}{40.92$\pm$2.25} & \multicolumn{1}{c|}{37.92$\pm$2.24} & \multicolumn{1}{c}{24.42$\pm$1.47} & \multicolumn{1}{c}{24.33$\pm$3.66} \\
    \midrule
    FedPer & \multicolumn{1}{c}{50.72$\pm$0.63} & \multicolumn{1}{c|}{53.87$\pm$0.58} & \multicolumn{1}{c}{41.80$\pm$2.15} & \multicolumn{1}{c|}{49.83$\pm$1.66} & \multicolumn{1}{c}{23.26$\pm$1.74} & \multicolumn{1}{c}{31.74$\pm$4.21} \\
    LG-FedAvg & \multicolumn{1}{c}{50.11$\pm$0.80} & \multicolumn{1}{c|}{51.80$\pm$0.67} & \multicolumn{1}{c}{41.49$\pm$2.56} & \multicolumn{1}{c|}{44.59$\pm$1.88} & \multicolumn{1}{c}{24.28$\pm$1.76} & \multicolumn{1}{c}{26.21$\pm$3.59} \\
    \midrule
    FedGen & 52.94$\pm$2.38 & 48.49$\pm$3.13 & 38.13$\pm$4.89 & 40.85$\pm$3.87 & 23.33$\pm$1.86 & 27.84$\pm$2.67 \\ 
    FedCG & 39.39$\pm$5.23 & \multicolumn{1}{c|}{37.06$\pm$4.35} & \multicolumn{1}{c}{30.44$\pm$3.30} & \multicolumn{1}{c|}{26.79$\pm$2.82} & \multicolumn{1}{c}{17.65$\pm$3.45} & \multicolumn{1}{c}{16.75$\pm$2.40} \\
    FedMD-CG & \textbf{54.82$\pm$0.79} & \textbf{55.18$\pm$1.75} & \textbf{46.30$\pm$2.24} & 47.56$\pm$2.21 & \textbf{26.18$\pm$1.79} & 30.55$\pm$2.42 \\
    \bottomrule
    \end{tabular}}%
  \label{tab:addlabel}%
\end{table*}%

\begin{figure*}[h]\captionsetup[subcaption]{font=scriptsize}
	\centering
	\begin{subfigure}{0.45\linewidth}
		\centering
		\includegraphics[width=1.0\linewidth]{CIFAR-10_10_comm_round.pdf}
		\caption{CIFAR-10, $\omega=10.0$}
		\label{fig3-a:}
	\end{subfigure}
	\centering
	\begin{subfigure}{0.45\linewidth}
		\centering
		\includegraphics[width=1.0\linewidth]{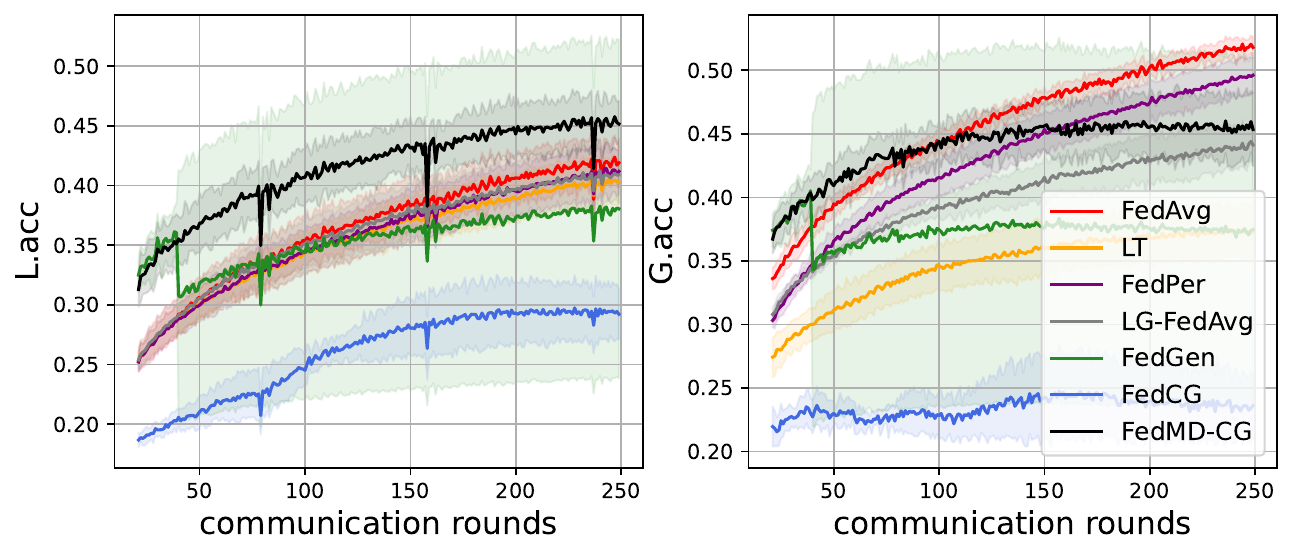}
		\caption{CIFAR-10, $\omega=1.0$}
		\label{fig3-b:}
	\end{subfigure}
        \centering
	\begin{subfigure}{0.45\linewidth}
		\centering
		\includegraphics[width=1.0\linewidth]{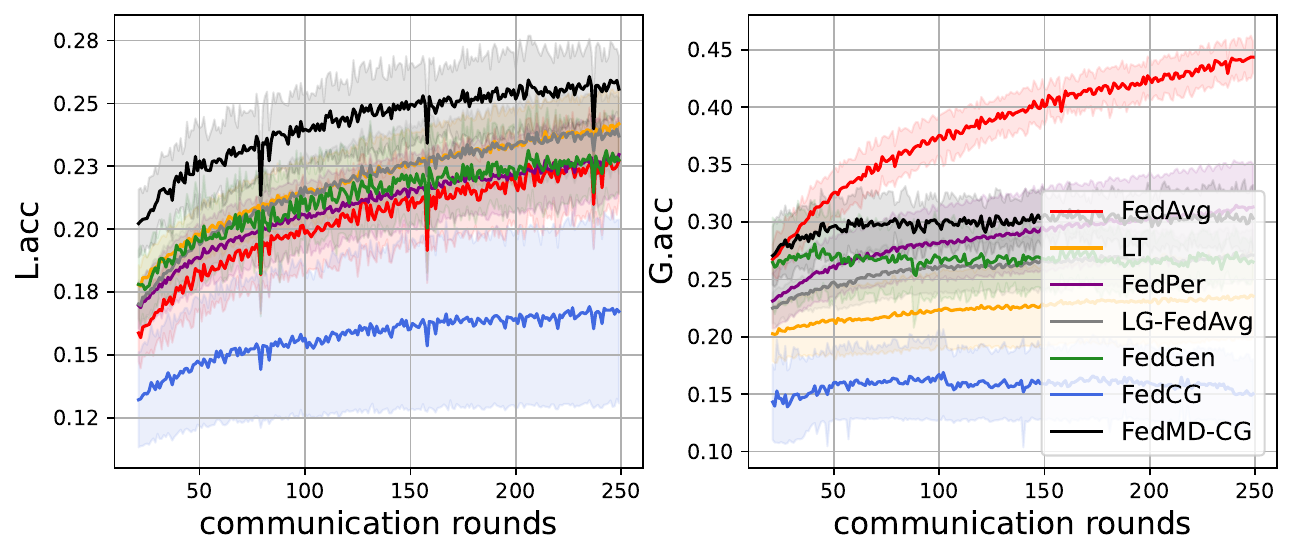}
		\caption{CIFAR-10, $\omega=0.1$}
		\label{fig3-c:}
	\end{subfigure}
	\caption{Learning curves for FedMD-CG as well as baselines over over CIFAR-10. } 
        \vspace*{-2ex}
\end{figure*}

\begin{table}[h]
  \centering
  \caption{Test performance~(\%) comparison between FedMD-CG and FedCG with different server-side aggregation manners over EMNIST. Note that \textit{AVE\_agg} and \textit{AVE\_agg$^\star$} denote weighted average aggregation.
  Specifically, FedMD-CG with \textit{AVE\_agg} transfers the knowledge from the global generator to local models at both the latent feature level and the logit level, whereas FedMD-CG with \textit{AVE\_agg$^\star$} transfers the knowledge from the global generator to local models only at the latent feature level.
  Also, \textit{KD\_agg} and \textit{KDC\_agg} denote the server-side aggregation manners from FedCG and FedMD-CG, respectively.}
    \resizebox{1.0\columnwidth}{!}{
    \begin{tabular}{c|l|ll|ll|ll}
    \toprule
    \multirow{3}[5]{*}{Alg.s} & \multicolumn{1}{c|}{\multirow{3}[5]{*}{Agg.}} & \multicolumn{6}{c}{EMNIST} \\
\cmidrule{3-8}          &       & \multicolumn{2}{c|}{$\omega=10.0$} & \multicolumn{2}{c|}{$\omega=1.0$} & \multicolumn{2}{c}{$\omega=0.1$} \\
\cmidrule{3-8}          &       & \multicolumn{1}{c}{$L.acc$} & \multicolumn{1}{c|}{$G.acc$} & \multicolumn{1}{c}{$L.acc$} & \multicolumn{1}{c|}{$G.acc$} & \multicolumn{1}{c}{$L.acc$} & \multicolumn{1}{c}{$G.acc$} \\
    \midrule
    \multirow{3}[2]{*}{FedCG} & \textit{AVE\_agg$^\star$}  & 97.26$\pm$0.02 & 98.00$\pm$0.04 & 95.05$\pm$0.17 & 97.05$\pm$0.24 & 50.55$\pm$4.32 & 86.74$\pm$1.44 \\
          & \textit{KD\_agg} & 97.67$\pm$0.03 & 98.08$\pm$0.07 & \textbf{96.06$\pm$0.33} & 97.70$\pm$0.16 & 49.91$\pm$3.83 & 87.66$\pm$2.08 \\
          & \textit{KDC\_agg} & 96.02$\pm$0.27 & 96.78$\pm$0.28 & 93.17$\pm$0.63 & 96.31$\pm$0.27 & 39.65$\pm$4.67 & 82.32$\pm$4.66 \\
    \midrule
    \multirow{4}[2]{*}{FedMD-CG} & \textit{AVE\_agg$^\star$}  & 97.29$\pm$0.01 & 98.19$\pm$0.03 & 95.12$\pm$0.46 & 97.21$\pm$0.21 & 51.36$\pm$3.63 & 86.88$\pm$1.53 \\
          & \textit{AVE\_agg}  & \textbf{97.74$\pm$0.05} & \textbf{98.36$\pm$0.04} & 95.51$\pm$0.25 & \textbf{97.86$\pm$0.08} & 52.62$\pm$3.74 & 86.95$\pm$1.35 \\
          & \textit{KD\_agg} & 96.69$\pm$0.11 & 97.19$\pm$0.10 & 94.70$\pm$0.44 & 96.72$\pm$0.22 & 53.14$\pm$4.73 & 83.86$\pm$2.10 \\
          & \textit{KDC\_agg} & 96.97$\pm$0.05 & 97.41$\pm$0.11 & 95.45$\pm$0.25 & 97.18$\pm$0.17 & \textbf{54.45$\pm$3.56} & \textbf{87.87$\pm$1.64} \\
    \bottomrule
    \end{tabular}}%
  \label{tab:addlabel}%
\end{table}%

\begin{figure}[h]\captionsetup[subfigure]{font=scriptsize}
    \centering
	\begin{subfigure}{0.32\linewidth}
		\centering
		\includegraphics[width=1.0\linewidth]{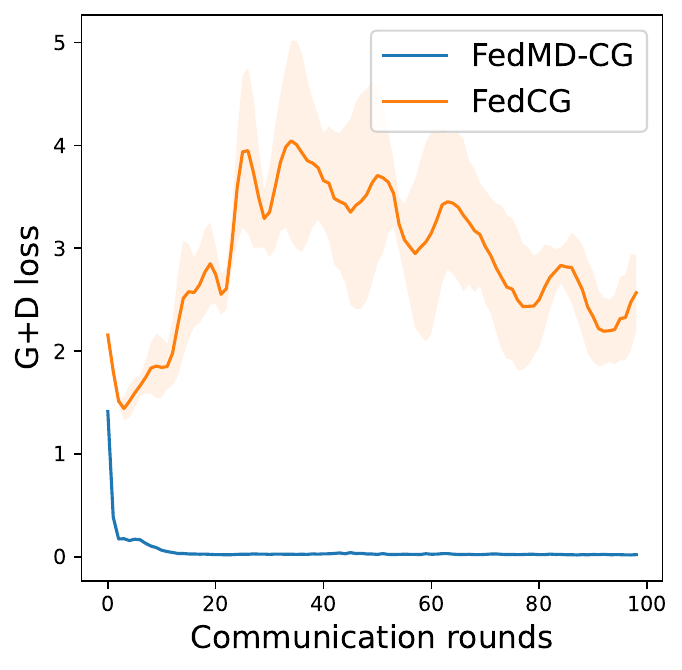}
		\caption{EMNIST, $\omega=0.1$}
		\label{chutian3}
	\end{subfigure}
	\centering
	\begin{subfigure}{0.32\linewidth}
		\centering
		\includegraphics[width=1.0\linewidth]{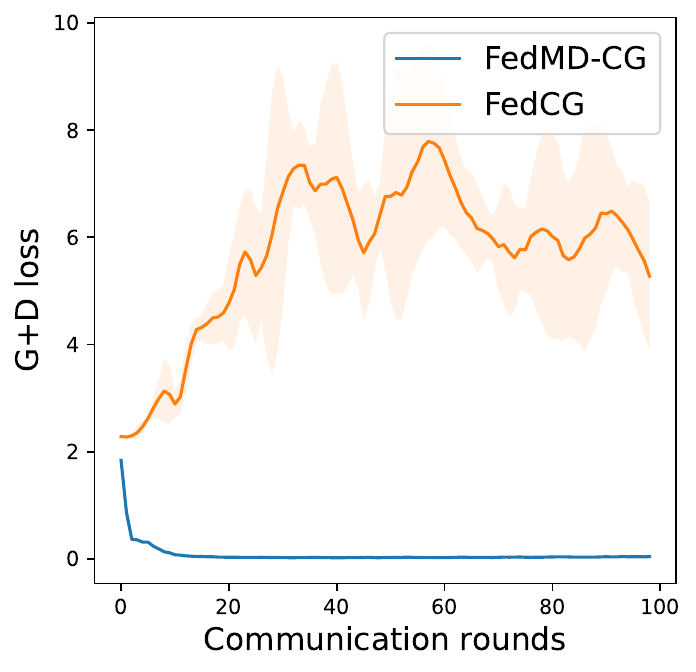}
		\caption{EMNIST, $\omega=1.0$}
		\label{chutian3}
	\end{subfigure}
	\centering
	\begin{subfigure}{0.32\linewidth}
		\centering
		\includegraphics[width=1.0\linewidth]{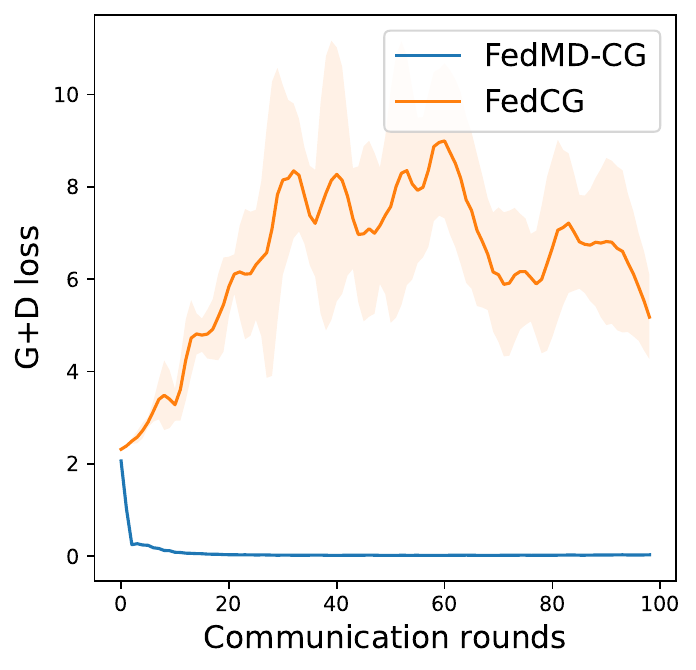}
		\caption{EMNIST, $\omega=10.0$}
		\label{chutian3}
	\end{subfigure}
	\caption{The consistency comparison between local generators and classifiers for FedCG and FedMD-CG w.r.t. \textit{AVE\_agg$^\star$} over EMNIST. G+D loss denotes the classification loss of the local classifier on the output of the local generator.} 
\end{figure}

\begin{table*}[h]
  \centering
  \caption{Test performance~(\%) comparison between FedMD-CG and FedCG with different server-side aggregation manners over FMNIST. Note that \textit{AVE\_agg} and \textit{AVE\_agg$^\star$} denote weighted average aggregation.
  Specifically, FedMD-CG with \textit{AVE\_agg} transfers the knowledge from the global generator to local models at both the latent feature level and the logit level, whereas FedMD-CG with \textit{AVE\_agg$^\star$} transfers the knowledge from the global generator to local models only at the latent feature level.
  Also, \textit{KD\_agg} and \textit{KDC\_agg} denote the server-side aggregation manners from FedCG and FedMD-CG, respectively.}
    \resizebox{1.0\columnwidth}{!}{
    \begin{tabular}{c|l|ll|ll|ll}
    \toprule
    \multirow{3}[5]{*}{Alg.s} & \multicolumn{1}{c|}{\multirow{3}[5]{*}{Agg.}} & \multicolumn{6}{c}{FMNIST} \\
\cmidrule{3-8}          &       & \multicolumn{2}{c|}{$\omega=10.0$} & \multicolumn{2}{c|}{$\omega=1.0$} & \multicolumn{2}{c}{$\omega=0.1$} \\
\cmidrule{3-8}          &       & \multicolumn{1}{c}{$L.acc$} & \multicolumn{1}{c|}{$G.acc$} & \multicolumn{1}{c}{$L.acc$} & \multicolumn{1}{c|}{$G.acc$} & \multicolumn{1}{c}{$L.acc$} & \multicolumn{1}{c}{$G.acc$} \\
    \midrule
    \multirow{3}[2]{*}{FedCG} & \textit{AVE\_agg$^\star$}  & 83.08$\pm$0.20 & 85.13$\pm$0.23 & 77.53$\pm$1.53 & 82.87$\pm$0.31 & 39.46$\pm$3.40 & 67.41$\pm$3.59 \\
          & \textit{KD\_agg} & 82.29$\pm$0.63 & 84.58$\pm$0.77 & 74.92$\pm$2.11 & 81.74$\pm$0.81 & 34.97$\pm$2.55 & 54.61$\pm$2.67 \\
          & \textit{KDC\_agg} & 83.28$\pm$3.49 & 85.52$\pm$3.37 & 75.64$\pm$2.26 & 82.63$\pm$0.48 & 37.23$\pm$2.54 & 62.89$\pm$7.01 \\
    \midrule
    \multirow{4}[2]{*}{FedMD-CG} & \textit{AVE\_agg$^\star$}  & 83.32$\pm$0.14 & 85.88$\pm$0.17 & 78.01$\pm$1.14 & 83.79$\pm$0.73 & 40.55$\pm$3.55 & 67.34$\pm$5.41 \\
          & \textit{AVE\_agg}  & 83.78$\pm$0.19 & 86.57$\pm$0.11 & 78.94$\pm$1.43 & \textbf{84.91$\pm$0.48} & 41.44$\pm$2.98 & 67.88$\pm$6.07 \\
          & \textit{KD\_agg} & 83.07$\pm$0.23 & 85.65$\pm$0.14 & 78.08$\pm$1.51 & 83.89$\pm$0.67 & 41.79$\pm$3.54 & 64.68$\pm$4.31 \\
          & \textit{KDC\_agg} & \textbf{84.32$\pm$0.18} & \textbf{87.18$\pm$0.13} & \textbf{79.00$\pm$1.43} & 84.47$\pm$0.38 & \textbf{42.55$\pm$3.68} & \textbf{71.09$\pm$1.01} \\
    \bottomrule
    \end{tabular}}%
  \label{tab:addlabel}%
\end{table*}%

\begin{figure*}[h]\captionsetup[subfigure]{font=scriptsize}
    \centering
	\begin{subfigure}{0.32\linewidth}
		\centering
		\includegraphics[width=1.0\linewidth]{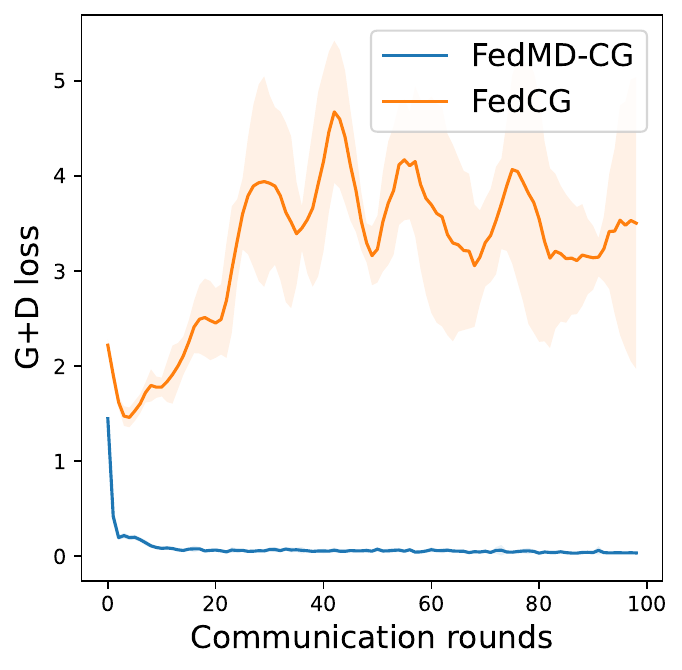}
		\caption{FMNIST, $\omega=0.1$}
		\label{chutian3}
	\end{subfigure}
	\centering
	\begin{subfigure}{0.32\linewidth}
		\centering
		\includegraphics[width=1.0\linewidth]{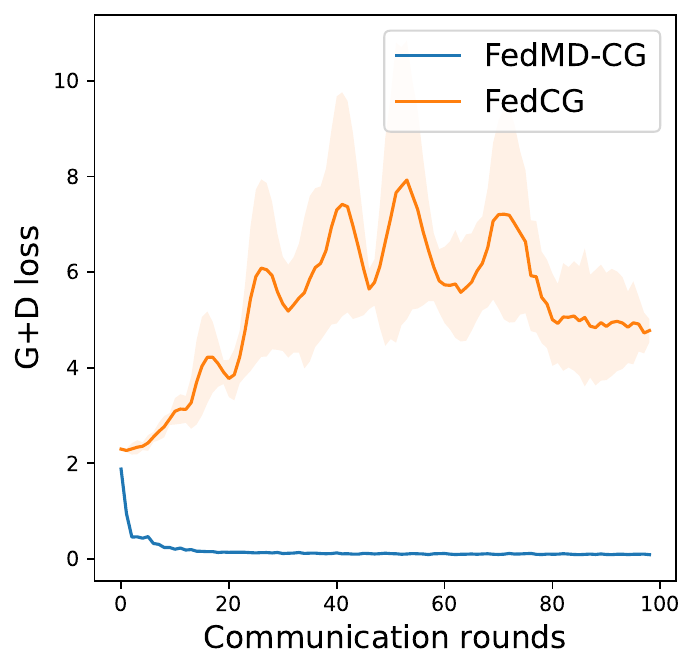}
		\caption{FMNIST, $\omega=1.0$}
		\label{chutian3}
	\end{subfigure}
	\centering
	\begin{subfigure}{0.32\linewidth}
		\centering
		\includegraphics[width=1.0\linewidth]{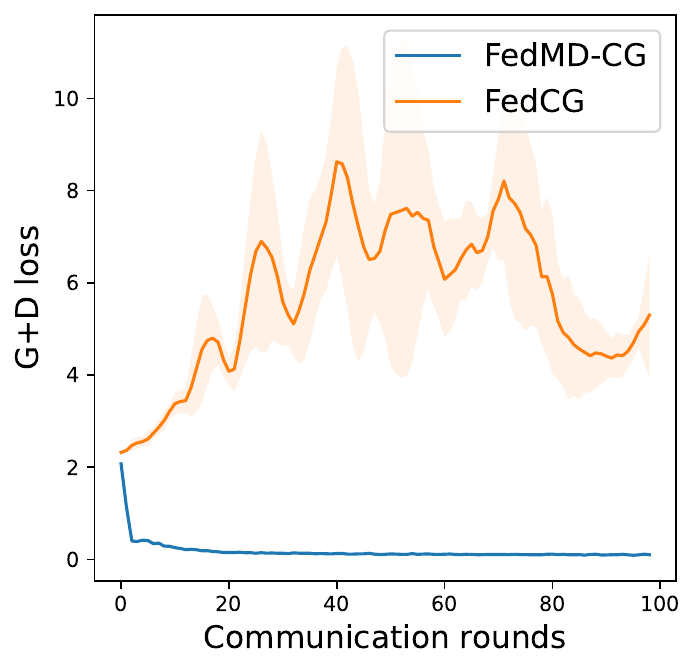}
		\caption{FMNIST, $\omega=10.0$}
		\label{chutian3}
	\end{subfigure}
	\caption{The consistency comparison between local generators and classifiers for FedCG and FedMD-CG w.r.t. \textit{AVE\_agg$^\star$} over FMNIST. G+D loss denotes the classification loss of the local classifier on the output of the local generator.} 
        \vspace*{-2ex}
\end{figure*}

\begin{table*}[h]
  \centering
  \caption{Test performance~(\%) comparison between FedMD-CG and FedCG with different server-side aggregation manners over CIFAR-10. Note that \textit{AVE\_agg} and \textit{AVE\_agg$^\star$} denote weighted average aggregation.
  Specifically, FedMD-CG with \textit{AVE\_agg} transfers the knowledge from the global generator to local models at both the latent feature level and the logit level, whereas FedMD-CG with \textit{AVE\_agg$^\star$} transfers the knowledge from the global generator to local models only at the latent feature level.
  Also, \textit{KD\_agg} and \textit{KDC\_agg} denote the server-side aggregation manners from FedCG and FedMD-CG, respectively.}
    \resizebox{1.0\columnwidth}{!}{
    \begin{tabular}{c|l|ll|ll|ll}
    \toprule
    \multirow{3}[6]{*}{Alg.s} & \multicolumn{1}{c|}{\multirow{3}[6]{*}{Agg.}} & \multicolumn{6}{c}{CIFAR-10} \\
\cmidrule{3-8}          &       & \multicolumn{2}{c|}{$\omega=10.0$} & \multicolumn{2}{c|}{$\omega=1.0$} & \multicolumn{2}{c}{$\omega=0.1$} \\
\cmidrule{3-8}          &       & \multicolumn{1}{c}{$L.acc$} & \multicolumn{1}{c|}{$G.acc$} & \multicolumn{1}{c}{$L.acc$} & \multicolumn{1}{c|}{$G.acc$} & \multicolumn{1}{c}{$L.acc$} & \multicolumn{1}{c}{$G.acc$} \\
    \midrule
    \multirow{3}[2]{*}{FedCG} & \textit{AVE\_agg$^\star$}  & 48.88$\pm$0.19 & 51.47$\pm$1.16 & 41.85$\pm$2.48 & 44.98$\pm$1.79 & 24.78$\pm$1.46 & 28.35$\pm$2.33 \\
          & \textit{KD\_agg} & \multicolumn{1}{c}{39.39$\pm$5.23} & \multicolumn{1}{c|}{37.06$\pm$4.35} & \multicolumn{1}{c}{30.44$\pm$3.30} & \multicolumn{1}{c|}{26.79$\pm$2.82} & \multicolumn{1}{c}{17.65$\pm$3.45} & \multicolumn{1}{c}{16.75$\pm$2.40} \\
          & \textit{KDC\_agg} & 28.70$\pm$2.44 & 29.14$\pm$0.54 & 28.87$\pm$0.90 & 25.92$\pm$1.53 & 21.63$\pm$2.02 & 26.44$\pm$3.91 \\
    \midrule
    \multirow{4}[2]{*}{FedMD-CG} & \textit{AVE\_agg$^\star$}  & 51.34$\pm$0.80 & 52.71$\pm$1.73 & 43.24$\pm$2.32 & 45.10$\pm$2.32 & 25.80$\pm$1.72 & 30.27$\pm$2.65 \\
          & \textit{AVE\_agg}  & 52.34$\pm$0.80 & 53.71$\pm$1.73 & 45.16$\pm$2.35 & 46.72$\pm$2.32 & 25.81$\pm$1.82 & 30.70$\pm$2.12 \\
          & \textit{KD\_agg} & 52.48$\pm$1.03 & 53.71$\pm$1.65 & 45.12$\pm$2.30 & 46.98$\pm$2.60 & 26.06$\pm$1.71 & \textbf{31.04$\pm$2.60} \\
          & \textit{KDC\_agg} & \textbf{54.82$\pm$0.79} & \textbf{55.18$\pm$1.75} & \textbf{46.30$\pm$2.24} & \textbf{47.56$\pm$2.21} & \textbf{26.18$\pm$1.79} & 30.55$\pm$2.42 \\
    \bottomrule
    \end{tabular}}%
  \label{tab:addlabel}%
\end{table*}%

\begin{figure*}[h]\captionsetup[subfigure]{font=scriptsize}
    \centering
	\begin{subfigure}{0.32\linewidth}
		\centering
		\includegraphics[width=1.0\linewidth]{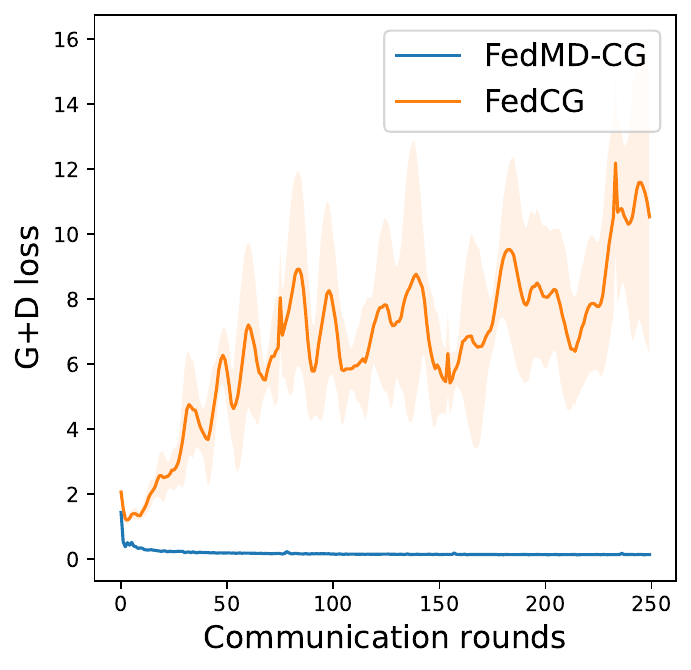}
		\caption{CIFAR-10, $\omega=0.1$}
		\label{chutian3}
	\end{subfigure}
	\centering
	\begin{subfigure}{0.32\linewidth}
		\centering
		\includegraphics[width=1.0\linewidth]{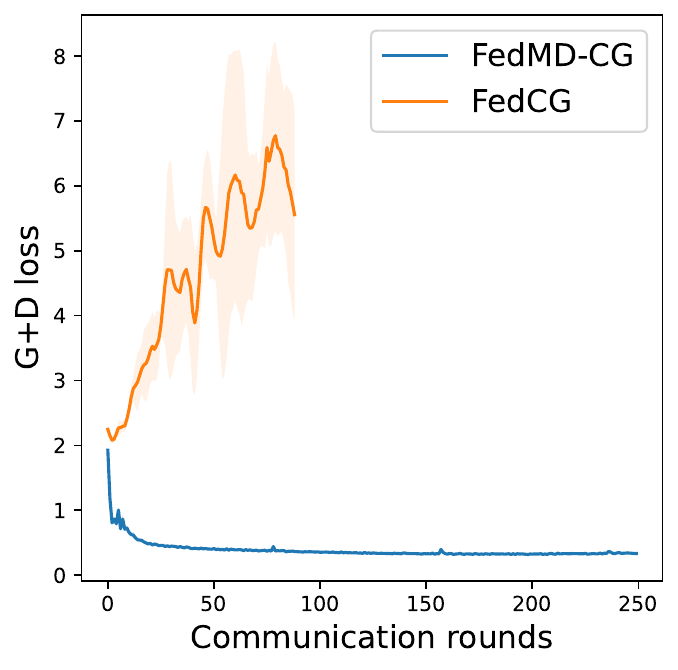}
		\caption{CIFAR-10, $\omega=1.0$}
		\label{chutian3}
	\end{subfigure}
	\centering
	\begin{subfigure}{0.32\linewidth}
		\centering
		\includegraphics[width=1.0\linewidth]{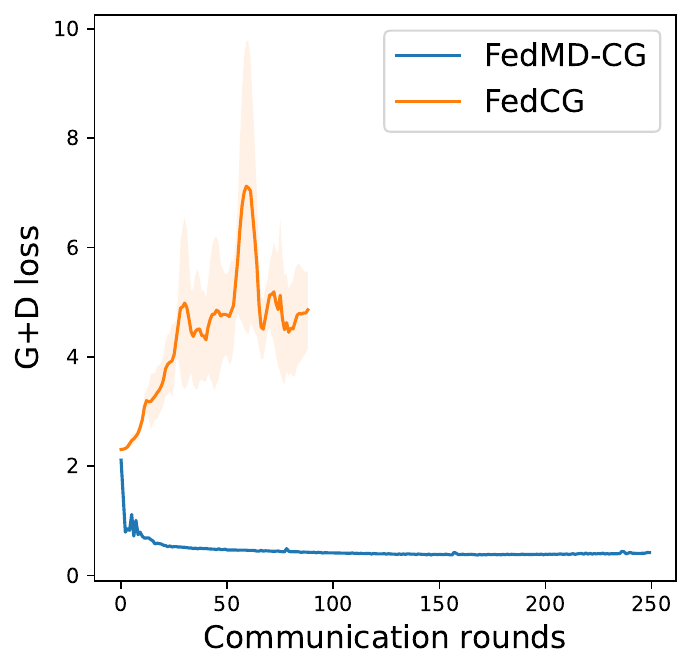}
		\caption{CIFAR-10, $\omega=10.0$}
		\label{chutian3}
	\end{subfigure}
	\caption{The consistency comparison between local generators and classifiers for FedCG and FedMD-CG w.r.t. \textit{AVE\_agg$^\star$} over CIFAR-10. G+D loss denotes the classification loss of the local classifier on the output of the local generator.} 
\end{figure*}

\begin{table}[htbp]
  \centering
  \caption{Impact of each loss for client-side training in FedMD-CG over EMNIST with $\omega=0.1$. Note that L.M.U and L.G.U denote the local model update and the local generator update, respectively. Also, we omit the subscript $i$ of each loss for client $i$.}
    \resizebox{1.0\columnwidth}{!}{
    \begin{tabular}{ccc|ccc}
    \toprule
    \multicolumn{6}{c}{FedMD-CG (baseline)} \\
    \midrule
    \multicolumn{3}{c}{$L.acc$} & \multicolumn{3}{c}{$G.acc$} \\
    \midrule
    \multicolumn{3}{c}{\textbf{54.45$\pm$3.56}} & \multicolumn{3}{c}{\textbf{87.87$\pm$1.64}} \\
    \midrule
    \midrule
    L.M.U & $L.acc$ & $G.acc$ & L.G.U & $L.acc$ & $G.acc$ \\
    \midrule
    $-{\mathop{\mathcal{L}}\limits ^{\to}}_{ce}$ & 52.51$\pm$3.61 & 85.44$\pm$1.40 & $-{\mathop{\mathcal{L}}\limits ^{\gets}}_{mse}$ & 49.33$\pm$3.27 & 81.09$\pm$4.49 \\
    $-{\mathop{\mathcal{L}}\limits ^{\to}}_{mse}$ & 52.85$\pm$3.17 & 86.47$\pm$1.54 & $-{\mathop{\mathcal{L}}\limits ^{\gets}}_{ce}$ & 51.87$\pm$2.67 & 82.97$\pm$1.54 \\
    $-{\mathop{\mathcal{L}}\limits ^{\to}}_{kl}$ & 52.31$\pm$2.29 & 84.70$\pm$1.03 & $-\mathcal{L}_{div}$ & 52.17$\pm$3.68 & 87.86$\pm$2.15 \\
    $-{\mathop{\mathcal{L}}\limits ^{\to}}_{ce}$, $-{\mathop{\mathcal{L}}\limits ^{\to}}_{mse}$ & 51.55$\pm$3.42 & 84.35$\pm$1.20 & $-{\mathop{\mathcal{L}}\limits ^{\gets}}_{mse}$, $-{\mathop{\mathcal{L}}\limits ^{\gets}}_{ce}$ & 47.81$\pm$2.95 & 80.66$\pm$3.45 \\
    $-{\mathop{\mathcal{L}}\limits ^{\to}}_{ce}$, $-{\mathop{\mathcal{L}}\limits ^{\to}}_{kl}$ & 50.42$\pm$3.43 & 82.59$\pm$1.37 & $-{\mathop{\mathcal{L}}\limits ^{\gets}}_{mse}$, $-\mathcal{L}_{div}$ & 47.12$\pm$2.65 & 80.49$\pm$2.05 \\
    $-{\mathop{\mathcal{L}}\limits ^{\to}}_{mse}$, $-{\mathop{\mathcal{L}}\limits ^{\to}}_{kl}$ & 51.34$\pm$3.16 & 85.63$\pm$1.07 & $-{\mathop{\mathcal{L}}\limits ^{\gets}}_{ce}$, $-\mathcal{L}_{div}$ & 50.32$\pm$3.35 & 82.09$\pm$2.38 \\
    $-{\mathop{\mathcal{L}}\limits ^{\to}}_{ce}$, $-{\mathop{\mathcal{L}}\limits ^{\to}}_{mse}$, $-{\mathop{\mathcal{L}}\limits ^{\to}}_{kl}$ & 46.53$\pm$4.77 & 84.51$\pm$2.24 & $-{\mathop{\mathcal{L}}\limits ^{\gets}}_{mse}$, $-{\mathop{\mathcal{L}}\limits ^{\gets}}_{ce}$, $-\mathcal{L}_{div}$ & 40.12$\pm$2.15 & 61.53$\pm$3.55 \\
    \bottomrule
    \end{tabular}}%
  \label{tab:addlabel}%
\end{table}%

\begin{table}[htbp]
  \centering
  \caption{Impact of each loss for client-side training in FedMD-CG over FMNIST with $\omega=1.0$. Note that L.M.U and L.G.U denote the local model update and the local generator update, respectively. Also, we omit the subscript $i$ of each loss for client $i$.}
    \resizebox{1.0\columnwidth}{!}{
    \begin{tabular}{ccc|ccc}
    \toprule
    \multicolumn{6}{c}{FedMD-CG (baseline)} \\
    \midrule
    \multicolumn{3}{c}{$L.acc$} & \multicolumn{3}{c}{$G.acc$} \\
    \midrule
    \multicolumn{3}{c}{\textbf{79.00$\pm$1.43}} & \multicolumn{3}{c}{\textbf{84.47$\pm$0.38}} \\
    \midrule
    \midrule
    L.M.U & $L.acc$ & $G.acc$ & L.G.U & $L.acc$ & $G.acc$ \\
    \midrule
    $-{\mathop{\mathcal{L}}\limits ^{\to}}_{ce}$ & 78.61$\pm$1.63 & 84.67$\pm$0.32 & $-{\mathop{\mathcal{L}}\limits ^{\gets}}_{mse}$ & 77.98$\pm$1.78 & 84.90$\pm$0.40 \\
    $-{\mathop{\mathcal{L}}\limits ^{\to}}_{mse}$ & 77.55$\pm$1.37 & 83.67$\pm$0.41 & $-{\mathop{\mathcal{L}}\limits ^{\gets}}_{ce}$ & 78.43$\pm$1.57 & 83.82$\pm$0.59 \\
    $-{\mathop{\mathcal{L}}\limits ^{\to}}_{kl}$ & 78.02$\pm$1.34 & 84.16$\pm$0.28 & $-\mathcal{L}_{div}$ & 77.87$\pm$1.49 & 83.84$\pm$0.41 \\
    $-{\mathop{\mathcal{L}}\limits ^{\to}}_{ce}$, $-{\mathop{\mathcal{L}}\limits ^{\to}}_{mse}$ & 77.02$\pm$1.79 & 82.73$\pm$0.24 & $-{\mathop{\mathcal{L}}\limits ^{\gets}}_{mse}$, $-{\mathop{\mathcal{L}}\limits ^{\gets}}_{ce}$ & 77.71$\pm$1.78 & 84.35$\pm$0.41 \\
    $-{\mathop{\mathcal{L}}\limits ^{\to}}_{ce}$, $-{\mathop{\mathcal{L}}\limits ^{\to}}_{kl}$ & 77.91$\pm$1.57 & 83.64$\pm$0.29 & $-{\mathop{\mathcal{L}}\limits ^{\gets}}_{mse}$, $-\mathcal{L}_{div}$ & 76.93$\pm$1.54 & 82.58$\pm$0.53 \\
    $-{\mathop{\mathcal{L}}\limits ^{\to}}_{mse}$, $-{\mathop{\mathcal{L}}\limits ^{\to}}_{kl}$ & 76.92$\pm$1.66 & 82.66$\pm$0.35 & $-{\mathop{\mathcal{L}}\limits ^{\gets}}_{ce}$, $-\mathcal{L}_{div}$ & 77.61$\pm$2.00 & 84.50$\pm$0.35 \\
    $-{\mathop{\mathcal{L}}\limits ^{\to}}_{ce}$, $-{\mathop{\mathcal{L}}\limits ^{\to}}_{mse}$, $-{\mathop{\mathcal{L}}\limits ^{\to}}_{kl}$ & 75.34$\pm$1.42 & 81.46$\pm$0.98 & $-{\mathop{\mathcal{L}}\limits ^{\gets}}_{mse}$, $-{\mathop{\mathcal{L}}\limits ^{\gets}}_{ce}$, $-\mathcal{L}_{div}$ & 72.26$\pm$1.85 & 79.41$\pm$0.15 \\
    \bottomrule
    \end{tabular}}%
  \label{tab:addlabel}%
\end{table}%

\begin{table}[htbp]
  \centering
  \caption{Impact of each loss for client-side training in FedMD-CG over CIFAR-10 with $\omega=10.0$. Note that L.M.U and L.G.U denote the local model update and the local generator update, respectively. Also, we omit the subscript $i$ of each loss for client $i$.}
    \resizebox{1.0\columnwidth}{!}{
    \begin{tabular}{ccc|ccc}
    \toprule
    \multicolumn{6}{c}{FedMD-CG (baseline)} \\
    \midrule
    \multicolumn{3}{c}{$L.acc$} & \multicolumn{3}{c}{$G.acc$} \\
    \midrule
    \multicolumn{3}{c}{\textbf{54.82$\pm$0.79}} & \multicolumn{3}{c}{\textbf{55.18$\pm$1.75}} \\
    \midrule
    \midrule
    L.M.U & $L.acc$ & $G.acc$ & L.G.U & $L.acc$ & $G.acc$ \\
    \midrule
    $-{\mathop{\mathcal{L}}\limits ^{\to}}_{ce}$ & 51.53$\pm$1.04 & 52.52$\pm$1.37 & $-{\mathop{\mathcal{L}}\limits ^{\gets}}_{mse}$ & 52.73$\pm$1.15 & 53.19$\pm$1.72 \\
    $-{\mathop{\mathcal{L}}\limits ^{\to}}_{mse}$ & 53.07$\pm$0.97 & 53.73$\pm$1.99 & $-{\mathop{\mathcal{L}}\limits ^{\gets}}_{ce}$ & 53.89$\pm$0.89 & 52.88$\pm$2.09 \\
    $-{\mathop{\mathcal{L}}\limits ^{\to}}_{kl}$ & 53.46$\pm$0.94 & 53.34$\pm$1.74 & $-\mathcal{L}_{div}$ & 52.66$\pm$0.77 & 53.11$\pm$1.72 \\
    $-{\mathop{\mathcal{L}}\limits ^{\to}}_{ce}$, $-{\mathop{\mathcal{L}}\limits ^{\to}}_{mse}$ & 51.02$\pm$0.52 & 52.96$\pm$1.01 & $-{\mathop{\mathcal{L}}\limits ^{\gets}}_{mse}$, $-{\mathop{\mathcal{L}}\limits ^{\gets}}_{ce}$ & 46.94$\pm$1.33 & 49.37$\pm$1.53 \\
    $-{\mathop{\mathcal{L}}\limits ^{\to}}_{ce}$, $-{\mathop{\mathcal{L}}\limits ^{\to}}_{kl}$ & 51.55$\pm$0.60 & 52.81$\pm$1.22 & $-{\mathop{\mathcal{L}}\limits ^{\gets}}_{mse}$, $-\mathcal{L}_{div}$ & 47.80$\pm$0.30 & 50.15$\pm$1.24 \\
    $-{\mathop{\mathcal{L}}\limits ^{\to}}_{mse}$, $-{\mathop{\mathcal{L}}\limits ^{\to}}_{kl}$ & 52.64$\pm$0.40 & 53.46$\pm$1.25 & $-{\mathop{\mathcal{L}}\limits ^{\gets}}_{ce}$, $-\mathcal{L}_{div}$ & 48.02$\pm$0.31 & 50.41$\pm$1.61 \\
    $-{\mathop{\mathcal{L}}\limits ^{\to}}_{ce}$, $-{\mathop{\mathcal{L}}\limits ^{\to}}_{mse}$, $-{\mathop{\mathcal{L}}\limits ^{\to}}_{kl}$ & 50.27$\pm$0.41 & 49.55$\pm$1.28 & $-{\mathop{\mathcal{L}}\limits ^{\gets}}_{mse}$, $-{\mathop{\mathcal{L}}\limits ^{\gets}}_{ce}$, $-\mathcal{L}_{div}$ & 44.33$\pm$1.38 & 47.66$\pm$1.35 \\
    \bottomrule
    \end{tabular}}%
  \label{tab:addlabel}%
\end{table}%

\begin{table*}[htbp]
  \centering
  \caption{Test performance~(\%) comparison among different diversity constraints used by FedMD-CG over EMNIST. Note that we omit the subscript $i$ of diversity
  loss for client $i$.}
    \resizebox{1.0\columnwidth}{!}{
    \begin{tabular}{c|ll|ll|ll}
    \toprule
    \multirow{3}[5]{*}{Div. con.} & \multicolumn{6}{c}{EMNIST} \\
\cmidrule{2-7}          & \multicolumn{2}{c|}{$\omega=10.0$} & \multicolumn{2}{c|}{$\omega=1.0$} & \multicolumn{2}{c}{$\omega=0.1$} \\
\cmidrule{2-7}          & \multicolumn{1}{c}{$L.acc$} & \multicolumn{1}{c|}{$G.acc$} & \multicolumn{1}{c}{$L.acc$} & \multicolumn{1}{c|}{$G.acc$} & \multicolumn{1}{c}{$L.acc$} & \multicolumn{1}{c}{$G.acc$} \\
    \midrule
    $\mathcal{L}_{div}^0$  & 96.96$\pm$0.06 & 97.41$\pm$0.11 & 95.39$\pm$0.21 & 97.17$\pm$0.17 & 53.09$\pm$4.27 & \textbf{88.85$\pm$1.31} \\
    $\mathcal{L}_{div}^1$  & \textbf{96.98$\pm$0.05} & 97.40$\pm$0.11 & 95.43$\pm$0.27 & 97.16$\pm$0.19 & 53.65$\pm$4.12 & 88.51$\pm$0.80 \\
    $\mathcal{L}_{div}^2$  & 96.97$\pm$0.05 & \textbf{97.41$\pm$0.11} & \textbf{95.45$\pm$0.25} & \textbf{97.18$\pm$0.17} & \textbf{54.45$\pm$3.56} & 87.87$\pm$1.64 \\
    \bottomrule
    \end{tabular}}%
  \label{tab:addlabel}%
\end{table*}%

\begin{table*}[htbp]
  \centering
  \caption{Test performance~(\%) comparison among different diversity constraints used by FedMD-CG over FMNIST. Note that we omit the subscript $i$ of diversity
  loss for client $i$.}
    \begin{tabular}{c|ll|ll|ll}
    \toprule
    \multirow{3}[6]{*}{Div. con.} & \multicolumn{6}{c}{FMNIST} \\
\cmidrule{2-7}          & \multicolumn{2}{c|}{$\omega=10.0$} & \multicolumn{2}{c|}{$\omega=1.0$} & \multicolumn{2}{c}{$\omega=0.1$} \\
\cmidrule{2-7}          & \multicolumn{1}{c}{$L.acc$} & \multicolumn{1}{c|}{$G.acc$} & \multicolumn{1}{c}{$L.acc$} & \multicolumn{1}{c|}{$G.acc$} & \multicolumn{1}{c}{$L.acc$} & \multicolumn{1}{c}{$G.acc$} \\
    \midrule
    $\mathcal{L}_{div}^0$  & 83.93$\pm$0.20 & \textbf{87.79$\pm$0.17} & 78.58$\pm$1.58 & 84.69$\pm$0.46 & 42.20$\pm$3.42 & 70.47$\pm$1.79 \\
    $\mathcal{L}_{div}^1$  & 84.10$\pm$0.16 & 87.60$\pm$0.12 & \textbf{79.03$\pm$1.52} & \textbf{84.73$\pm$0.49} & 42.25$\pm$3.36 & 70.28$\pm$1.26 \\
    $\mathcal{L}_{div}^2$  & \textbf{84.32$\pm$0.18} & 87.18$\pm$0.13 & 79.00$\pm$1.43 & 84.47$\pm$0.38 & \textbf{42.55$\pm$3.68} & \textbf{71.09$\pm$1.01} \\
    \bottomrule
    \end{tabular}%
  \label{tab:addlabel}%
\end{table*}%

\begin{table*}[htbp]
  \centering
  \caption{Test performance~(\%) comparison among different diversity constraints used by FedMD-CG over CIFAR-10. Note that we omit the subscript $i$ of diversity
  loss for client $i$.}
    \begin{tabular}{c|ll|ll|ll}
    \toprule
    \multirow{3}[6]{*}{Div. con.} & \multicolumn{6}{c}{CIFAR-10} \\
\cmidrule{2-7}          & \multicolumn{2}{c|}{$\omega=10.0$} & \multicolumn{2}{c|}{$\omega=1.0$} & \multicolumn{2}{c}{$\omega=0.1$} \\
\cmidrule{2-7}          & \multicolumn{1}{c}{$L.acc$} & \multicolumn{1}{c|}{$G.acc$} & \multicolumn{1}{c}{$L.acc$} & \multicolumn{1}{c|}{$G.acc$} & \multicolumn{1}{c}{$L.acc$} & \multicolumn{1}{c}{$G.acc$} \\
    \midrule
    $\mathcal{L}_{div}^0$  & \multicolumn{1}{c}{54.24$\pm$0.72} & \multicolumn{1}{c|}{54.78$\pm$1.88} & \multicolumn{1}{c}{46.36$\pm$2.36} & \multicolumn{1}{c|}{47.20$\pm$2.15} & \multicolumn{1}{c}{26.04$\pm$1.76} & \multicolumn{1}{c}{30.62$\pm$2.56} \\
    $\mathcal{L}_{div}^1$  & 54.81$\pm$0.71 & 54.90$\pm$1.73 & \textbf{46.38$\pm$2.37} & 47.53$\pm$2.14 & 26.14$\pm$1.80 & \textbf{30.70$\pm$2.55} \\
    $\mathcal{L}_{div}^2$  & \textbf{54.82$\pm$0.79} & \textbf{55.18$\pm$1.75} & 46.30$\pm$2.24 & \textbf{47.56$\pm$2.21} & \textbf{26.18$\pm$1.79} & 30.55$\pm$2.42 \\
    \bottomrule
    \end{tabular}%
  \label{tab:addlabel}%
\end{table*}%

\begin{figure*}[t]\captionsetup[subfigure]{font=scriptsize}
    \centering
    \begin{subfigure}{0.329\linewidth}
        \centering
        \includegraphics[width=1.0\linewidth]{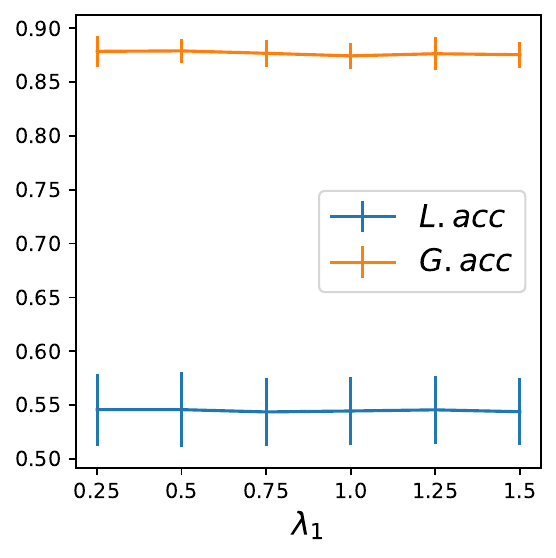}
        \label{chutian3}
    \end{subfigure}
    \centering
    \begin{subfigure}{0.329\linewidth}
        \centering
        \includegraphics[width=1.0\linewidth]{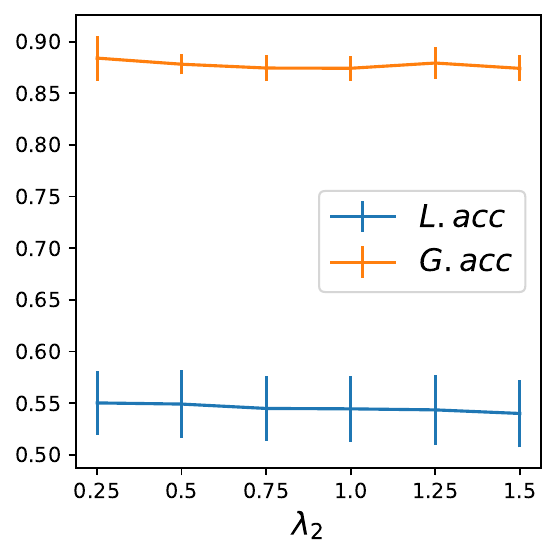}
        \label{chutian3}
    \end{subfigure}
    \centering
    \begin{subfigure}{0.329\linewidth}
        \centering
        \includegraphics[width=1.0\linewidth]{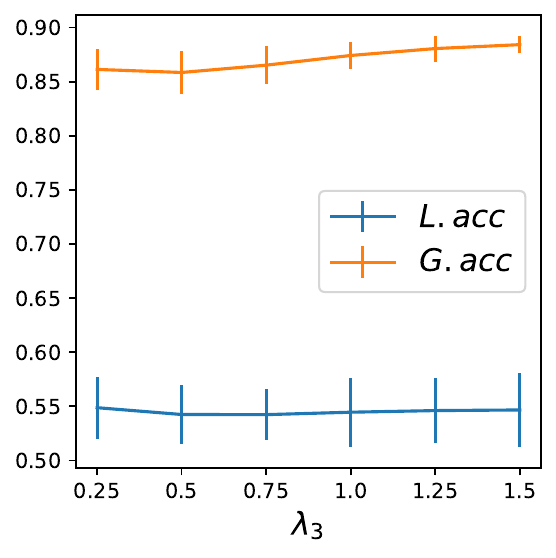}
        \label{chutian3}
    \end{subfigure} 
    \\
    \centering 
    \begin{subfigure}{0.329\linewidth}
        \centering
        \includegraphics[width=1.0\linewidth]{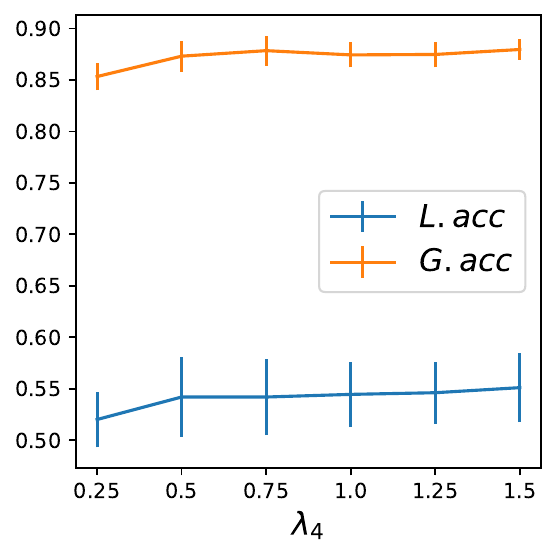}
        \label{chutian3}
    \end{subfigure}
    \centering
    \begin{subfigure}{0.329\linewidth}
        \centering
        \includegraphics[width=1.0\linewidth]{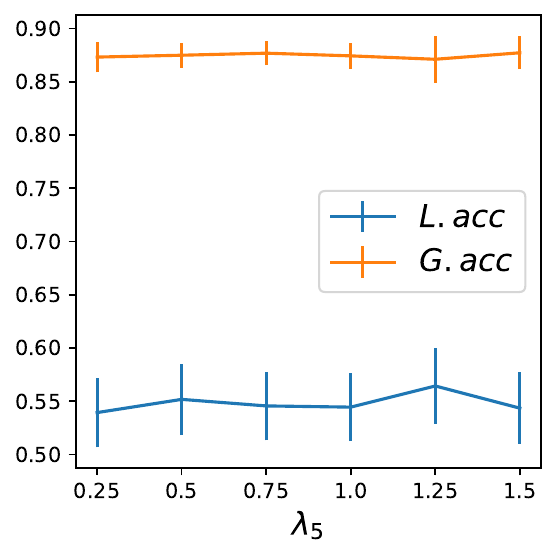}
        \label{chutian3}
    \end{subfigure}
    \centering
    \begin{subfigure}{0.329\linewidth}
        \centering
        \includegraphics[width=1.0\linewidth]{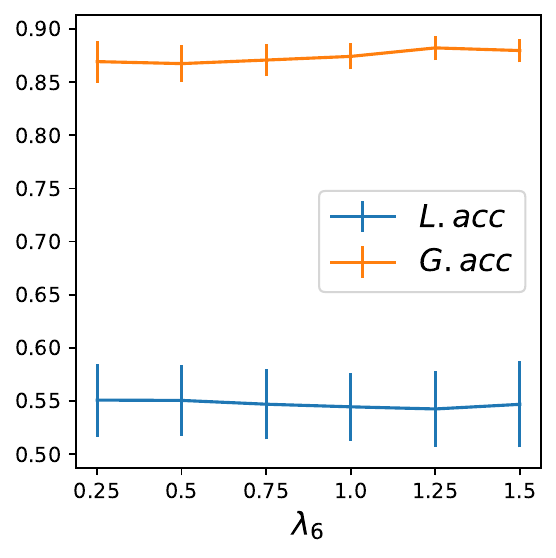}
        \label{chutian3}
    \end{subfigure}
    \caption{Test performance of FedMD-CG using varying hyperparameters on EMNIST with $\omega=0.1$} 
\end{figure*}

\begin{figure*}[t]\captionsetup[subfigure]{font=scriptsize}
    \centering
    \begin{subfigure}{0.329\linewidth}
        \centering
        \includegraphics[width=1.0\linewidth]{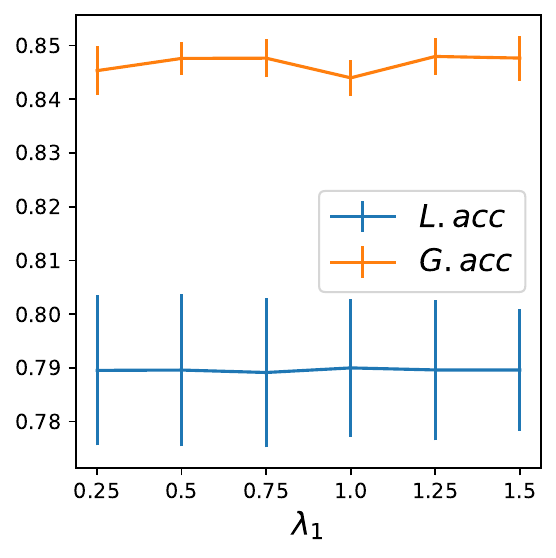}
        \caption{$\lambda_{1}$}
        \label{chutian3}
    \end{subfigure}
    \centering
    \begin{subfigure}{0.329\linewidth}
        \centering
        \includegraphics[width=1.0\linewidth]{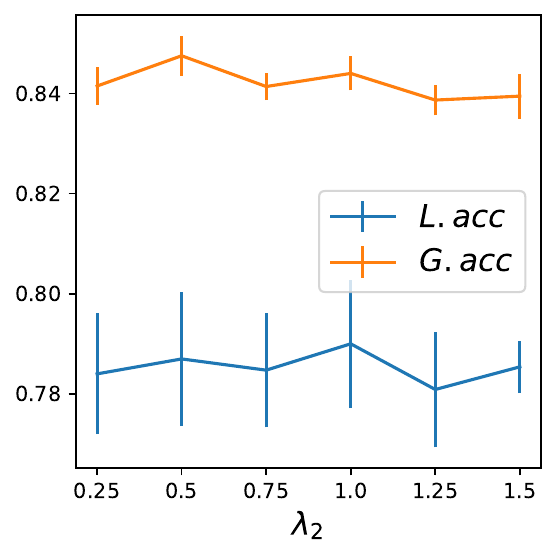}
        \caption{$\lambda_{2}$}
        \label{chutian3}
    \end{subfigure}
    \centering
    \begin{subfigure}{0.329\linewidth}
        \centering
        \includegraphics[width=1.0\linewidth]{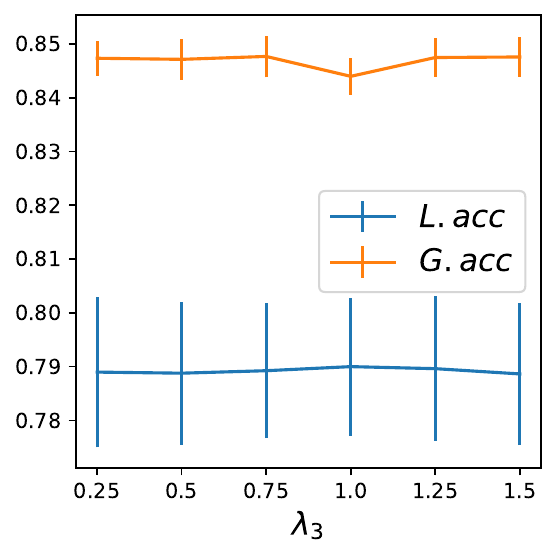}
        \caption{$\lambda_{3}$}
        \label{chutian3}
    \end{subfigure} \\
    \centering
    \begin{subfigure}{0.329\linewidth}
        \centering
        \includegraphics[width=1.0\linewidth]{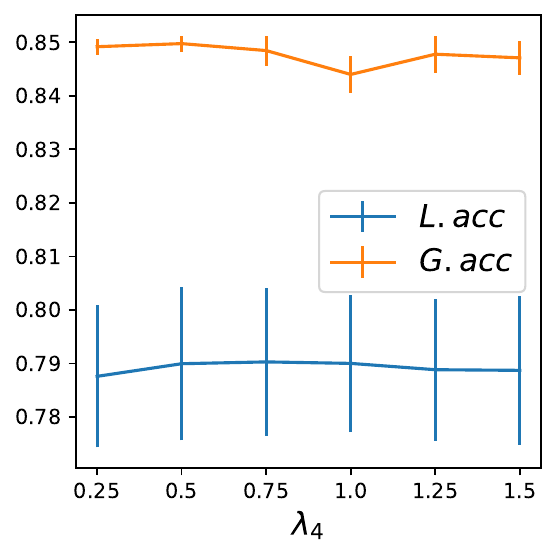}
        \caption{$\lambda_{4}$}
        \label{chutian3}
    \end{subfigure}
    \centering
    \begin{subfigure}{0.329\linewidth}
        \centering
        \includegraphics[width=1.0\linewidth]{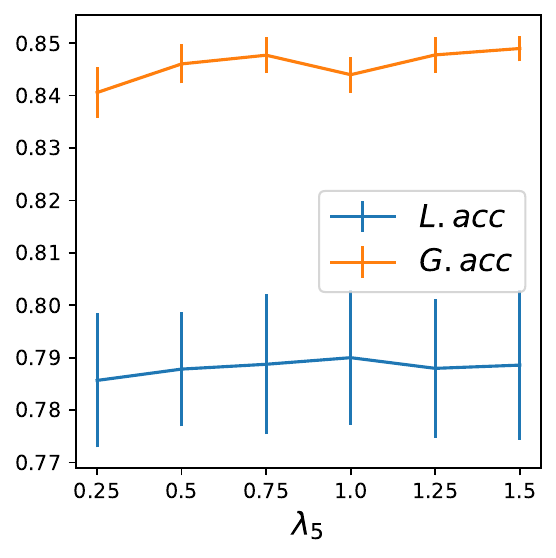}
        \caption{$\lambda_{5}$}
        \label{chutian3}
    \end{subfigure}
    \centering
    \begin{subfigure}{0.329\linewidth}
        \centering
        \includegraphics[width=1.0\linewidth]{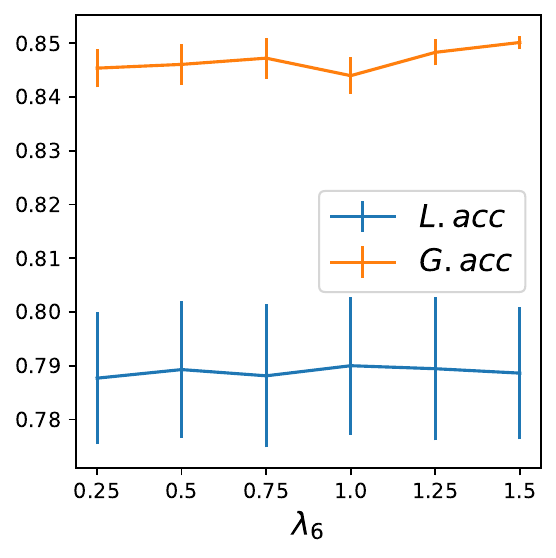}
        \caption{$\lambda_{6}$}
        \label{chutian3}
    \end{subfigure}
    \caption{Test performance of FedMD-CG using varying hyperparameters on FMNIST with $\omega=1.0$.} 
\end{figure*}

\begin{figure*}[t]\captionsetup[subfigure]{font=scriptsize}
    \centering
    \begin{subfigure}{0.329\linewidth}
        \centering
        \includegraphics[width=1.0\linewidth]{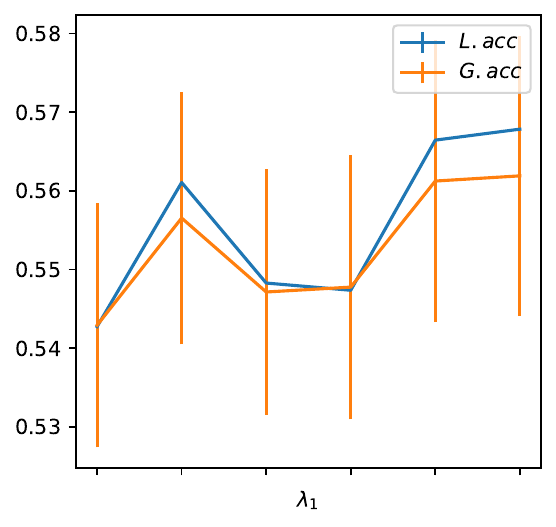}
        \label{chutian3}
    \end{subfigure}
    \centering
    \begin{subfigure}{0.329\linewidth}
        \centering
        \includegraphics[width=1.0\linewidth]{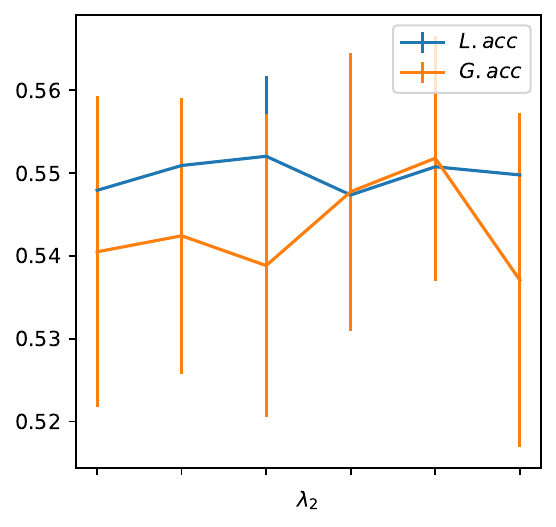}
        \label{chutian3}
    \end{subfigure}
    \centering
    \begin{subfigure}{0.329\linewidth}
        \centering
        \includegraphics[width=1.0\linewidth]{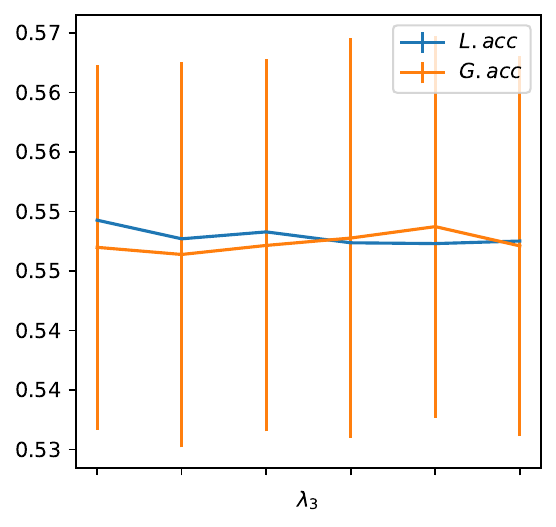}
        \label{chutian3}
    \end{subfigure} \\
    \centering
    \begin{subfigure}{0.329\linewidth}
        \centering
        \includegraphics[width=1.0\linewidth]{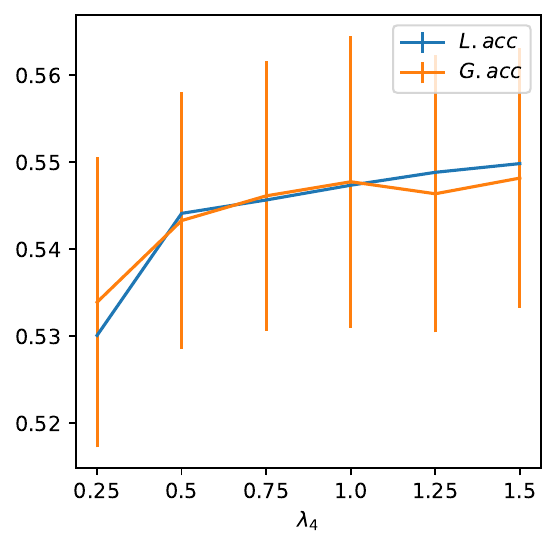}
        \label{chutian3}
    \end{subfigure}
    \centering
    \begin{subfigure}{0.329\linewidth}
        \centering
        \includegraphics[width=1.0\linewidth]{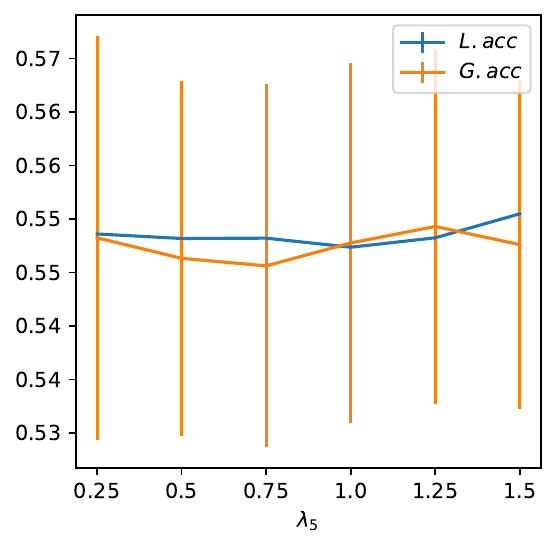}
        \label{chutian3}
    \end{subfigure}
    \centering
    \begin{subfigure}{0.329\linewidth}
        \centering
        \includegraphics[width=1.0\linewidth]{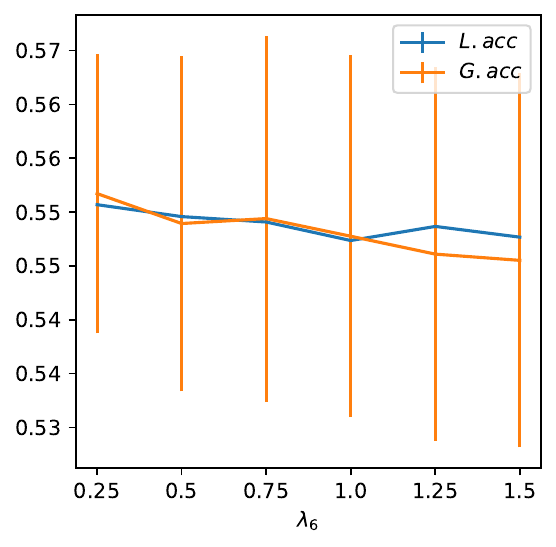}
        \label{chutian3}
    \end{subfigure}
    \caption{Test performance of FedMD-CG using varying hyperparameters on CIFAR-10 with $\omega=10.0$} 
\end{figure*}

\section{Limitations}
\label{app_discussion:}
In the field of Federated Learning~(FL), there are many trade-offs, including utility, privacy protection, computational efficiency and communication cost, etc.
It is well known that trying to develop a universal FL method that can address all problems is extremely challenging. 
In this work, we work on improving privacy leakage defects in FL while maintaining robust model performance.
Next, we discuss some of the limitations of FedMD-CG.

\textbf{Computational Efficiency, Communication Cost and Utility.} 
We acknowledge that deploying FedMD-CG in a real-world FL application requires clients to have more hardware and computational resources to train generators and local models as compared to FedAvg.
Specifically, compared with FedAvg or MD-based methods~(e.g., FedPer and LG-FedAvg), the training time of FedMD-CG will be longer, as it needs to additionally train the generator on the clients.
In our experiments, FedMD-CG takes two to three times longer to run per communication round than they do.
Moreover, compared to FedAvg or MD-based methods, FedMD-CG requires an additional vector of label statistics to be transmitted~(see line 18 in Algorithm~\ref{alg:1}).
However, the communication cost of this vector is negligible compared to that of the model.
We also acknowledge that FedMD-CG still has room for improvement in model performance.
Table~\ref{table1:} shows that FedAvg achieves the optimal model performance in many scenarios, so it is an attractive topic in the field of FL to achieve comparable test performance levels to FedAvg with high-level privacy protection.
Meanwhile, there is a trade-off between the capacity of the generator and the communication cost.

\textbf{Privacy Protection.}
Since FedMD-CG trains a local generator on each client for replacing the local feature extractor~(LFE) by simulating the output vector space of LFE, i.e., the latent feature space, rather than the distribution space of private data, it provides high-level privacy protection. 
In addition, FedMD-CG requires clients to upload the label statistics of the data, which also is at risk of compromising privacy.

\section{Broader Impacts}
\label{Broader_Impacts:}
We work on how to improve privacy leakage defects in FL while maintaining robust model performance.
Our work points out the pitfalls of the existing method FedCG.
First, knowledge transfer modality at the latent feature level may not be sufficient. 
Second, additional discriminators need to be trained to satisfy the adversarial training of cGAN. 
Third, the trained local generator may not match the local classifier, terming their inconsistency.
Our proposed FedMD-CG can deal with the said issues well.
FedMD-CG exemplifies potential positive impacts on society, enabling models with superior performance while ensuring high-level privacy protection in real-world FL applications.
Meanwhile, FedMD-CG may have negative social impacts related to high resource consumption.
FedMD-CG-based FL systems require more client-side power resources to train the generator and local model.
FedMD-CG does not involve social ethics.


\end{document}